\documentclass[runningheads]{llncs}
\usepackage{graphicx}
\usepackage{amsmath,amssymb} 
\usepackage{color}
\usepackage{tabularx}
\usepackage{booktabs}
\usepackage[hyphens]{url}
\graphicspath{{fig/}}
\usepackage{epsfig}
\usepackage{graphicx}
\usepackage{subcaption}
\usepackage{mathtools}
\usepackage{footmisc}

\usepackage[utf8]{inputenc}
\usepackage[T1]{fontenc}
\usepackage[dvipsnames]{xcolor}
\usepackage{algorithm}
\usepackage[noend]{algpseudocode}
\usepackage{graphicx}
\usepackage{booktabs}
\usepackage{multirow}
\usepackage{multicol}
\usepackage{xspace}
\usepackage{float}

\iftrue 

\newcommand{\cutsectionup}{\vspace*{0in}}
\newcommand{\cutsectiondown}{\vspace*{-0.05in}}

\newcommand{\cutsubsectionup}{\vspace*{-0.05in}}
\newcommand{\cutsubsectiondown}{\vspace*{-0.04in}}

\else 
\newcommand{\cutsectionup}{}
\newcommand{\cutsectiondown}{}

\newcommand{\cutsubsectionup}{}
\newcommand{\cutsubsectiondown}{}

\fi

\newcommand{\ba}{\mathbf{a}}
\newcommand{\bA}{\mathbf{A}}

\newcommand{\bx}{\mathbf{x}}
\newcommand{\bX}{\mathbf{X}}
\newcommand{\bI}{\mathbf{I}}

\newif\ifdraft

\newcommand{\ms}[1]{\ifdraft {\color{green}{#1}} \else {#1}\fi}
\newcommand{\kn}[1]{\ifdraft {\color{blue}{#1}} \else {#1}\fi}

\newcommand{\MS}[1]{\ifdraft {\color{green}{\textbf{MS: #1}}}\else {}\fi}
\newcommand{\KN}[1]{\ifdraft {\color{Fuchsia}{\textbf{KN: #1}}}\else {}\fi}

\newcommand{\comment}[1]{}
\newcommand{\Introfig}[1]{./images/#1}
\newcommand{\Analysisfig}[1]{./images/#1}

\newcommand{\Supfig}[1]{./figs/#1}

\begin{document}
	
	\pagestyle{headings}
	\mainmatter
	\def\ECCVSubNumber{}  
	\title{Towards Robust Fine-grained Recognition by Maximal Separation of Discriminative Features}
	\author{Krishna Kanth Nakka and Mathieu Salzmann \\ }
	\institute{CVLab, EPFL}	
	\maketitle


\begin{abstract}

Adversarial attacks have been widely studied for general classification tasks, but remain unexplored in the context of fine-grained recognition, where the inter-class similarities facilitate the attacker's task.
In this paper, we identify the proximity of the latent representations of different classes in fine-grained recognition networks as a key factor to the success of adversarial attacks. We therefore introduce an attention-based regularization mechanism that maximally separates the discriminative latent features of different classes while minimizing the contribution of the non-discriminative regions to the final class prediction. As evidenced by our experiments, this allows us to significantly improve robustness to adversarial attacks, to the point of matching or even surpassing that of adversarial training, but without requiring access to adversarial samples.


\end{abstract}
\cutsectionup
\section{Introduction}
\cutsectiondown
Deep networks yield impressive results in many computer vision tasks~\cite{long2015fully,krizhevsky2012imagenet,karpathy2014large,zhang2015cross}. Nevertheless, their performance degrades under adversarial attacks, where natural examples are perturbed with human-imperceptible, carefully crafted noise~\cite{fgsm}.  Adversarial attacks have been extensively studied for the task of general object recognition~\cite{JSMA,fgsm,bim,deepfool,carlini2017towards,dong2018boosting}, with much effort dedicated to studying and improving the robustness of deep networks to such attacks~\cite{madry2017towards,kannan2018adversarial,xie2019feature}. However, adversarial attacks and defense mechanisms for fine-grained recognition problems, where one can expect the inter-class similarities to facilitate the attacker's task, remain unexplored.


\providecommand{\localwidth}{}
\renewcommand{\localwidth}{\linewidth}

\providecommand{\localheight}{}
\renewcommand{\localheight}{3cm}

\begin{figure}[t]
	\footnotesize
	\centering
      \includegraphics[width=\textwidth]{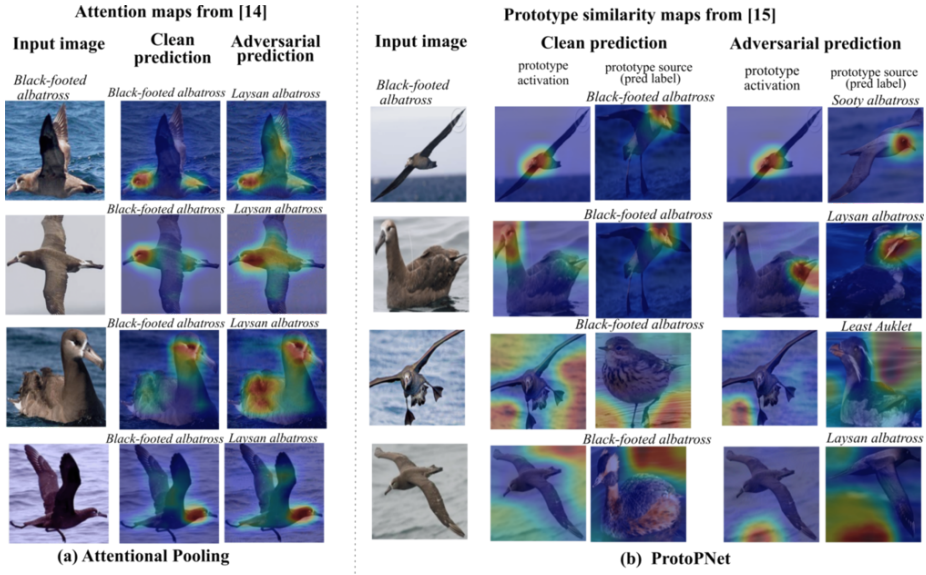}
	\caption{\textbf{Interpreting adversarial attacks for fine-grained recognition.} \small{We analyze the attention maps, obtained with~\cite{girdhar2017attentional}(a) and~\cite{chen2019looks}(b), of four images from the \emph{Black-footed albatross} class. Under attack, these images are misclassified as 
	closely-related bird species, such as \emph{Layman albatross}, because the classifiers focus on either confusing regions that look similar in these classes, such as the bird's beak, or non-discriminative background regions, such as water.}}
	\label{fig:analysis}
\end{figure}

In this paper, we therefore analyze the reasons for the success of adversarial attacks on fine-grained recognition techniques and introduce a defense mechanism to improve a network's robustness. To this end, we visualize the image regions mostly responsible for the classification results. Specifically, we consider both the attention-based framework of~\cite{girdhar2017attentional}, closely related to class activation maps (CAMs)~\cite{zhou2016learning}, and the recent prototypical part network (ProtoPNet) of~\cite{chen2019looks}, designed for fine-grained recognition, which relates local image regions to interpretable prototypes. As shown in Fig.~\ref{fig:analysis}, an adversarial example activates 
either confusing regions that look similar in samples from the true class and from the class activated by the adversarial attack, such as the beak of the bird, or, in the ProtoPNet case, non-discriminative background regions, such as water. 
thus making the network more vulnerable to attacks.


\providecommand{\localwidth}{}
\renewcommand{\localwidth}{\linewidth}

\providecommand{\localheight}{}
\renewcommand{\localheight}{5cm}

\begin{figure}[t]
	\footnotesize
	\centering
	\begin{tabular}{ccc}
		\includegraphics[height= \localheight,width=\localwidth]{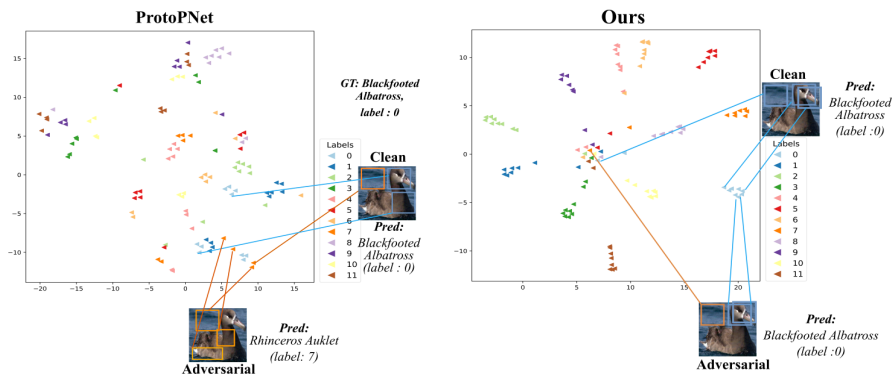}\\

	\end{tabular}
	\vspace{-3mm}
	\caption{{\bf t-SNE visualization of the prototypes from 12 fine-grained classes of the CUB200 dataset.}  In ProtoPNet~\cite{chen2019looks}, the prototypes of different classes are not well separated, making the network vulnerable to attacks. By contrast, our approach yields well-separated discriminative prototypes, while clustering the background ones, which,
by means of an attention mechanism
do not participate the prediction. This complicates the attacker's task. }
	\label{fig:teaser_tsne2}
\end{figure}

Motivated this observation, we introduce a defense mechanism based on the intuition that the discriminative regions of each class should be maximally separated from that of the other classes. To this end, we design an attention-aware model that pushes away the discriminative prototypes of the different classes. The effectiveness of our approach is illustrated in Fig.~\ref{fig:teaser_tsne2}, where the prototypes of different classes are nicely separated, except for those corresponding to non-discriminative regions.
However, by means of an attention mechanism, we enforce these non-discriminative prototypes to play no role in the final class prediction. Ultimately, our approach reduces the influence of the non-discriminative regions on the classification while increasing the magnitude of the displacement in the latent space that the attacker must perform to successfully move the network's prediction away from the true label.   As evidenced by our experiments, our approach significantly outperforms in robustness the baseline ProtoPNet and attentional pooling network, in some cases reaching adversarial accuracies on par with or higher than their adversarially-trained~\cite{tsipras2018robustness,tramer2017ensemble} counterparts, but at virtually no additional computational cost. 

Our main contributions can be summarized as follows.  We analyze and explain the decisions of fine-grained recognition networks by studying the image regions responsible for classification for both clean and adversarial examples. We design an interpretable, attention-aware network for robust fine-grained recognition by constraining the latent space of discriminative regions.  Our method improves robustness to a level comparable to that of 
adversarial training, without requiring access to adversarial samples and without trading off clean accuracy.  \ms{We will make our code publicly available upon acceptance of the paper.}


\section{Related Work}

\noindent{\textbf{Adversarial Robustness.} DNNs were first shown to be vulnerable to adversarial, human-imperceptible perturbations in the context of general image recognition. Such attacks were initially studied in~\cite{szegedy2013intriguing}, quickly followed by the simple single-step Fast Gradient Sign Method (FGSM)~\cite{fgsm} and its multiple-step BIM variant~\cite{bim}. In~\cite{dong2018boosting}, the attacks were stabilized by incorporating momentum in the gradient computation. Other popular attacks include DeepFool~\cite{deepfool}, which iteratively linearizes the classifier to compute minimal perturbations sufficient for the sample to cross the decision boundary, and other computationally more expensive attacks, such as CW~\cite{carlini2017towards}, JSMA~\cite{JSMA},  and others~\cite{xu2018structured,su2019one,narodytska2017simple}. As of today, Projected Gradient Descent (PGD)~\cite{madry2017towards}, which utilizes the local first-order network information to compute a maximum loss increment within a specified $\ell_{\infty}$ norm-bound, is generally considered as the most effective attack strategy.

Despite a significant research effort in devising defense mechanisms against adversarial attacks~\cite{papernot2016distillation,samangouei2018defense,xie2017mitigating,song2017pixeldefend}, it was shown in~\cite{athalye2018obfuscated} that most such defenses can easily be breached in the white-box setting, where the attacker knows the network architecture. The main exception to this rule is adversarial training~\cite{madry2017towards}, where the model is trained jointly with clean images and their adversarial counterparts. Many variants of adversarial training were thus proposed, such as ensemble adversarial training~\cite{tramer2017ensemble} to soften the classifier's decision boundaries,  ALP~\cite{kannan2018adversarial} to minimize the  difference between the logit activations of real and adversarial images, the use of additional feature denoising blocks~\cite{xie2019feature}, of metric learning~\cite{mao2019metric}, and of regularizers to penalize changes in the model's prediction w.r.t. the input perturbations~\cite{ross2018improving,zhang2019theoretically}.
Nevertheless, PGD-based adversarial training remains the method of choice, thanks to its robustness and generalizability to unseen attacks~\cite{fgsm,bim,deepfool,carlini2017towards}. 

Unfortunately, adversarial training is computationally expensive. This was tackled in~\cite{shafahi2019adversarial} by recycling the gradients computed to update the model parameters so as to reduce the overhead of generating adversarial examples, albeit not remove this overhead entirely.
More recently,~\cite{wong2020fast} showed that combining the single-step FGSM with random initialization is almost as effective as PGD-based training, but at a significantly lower cost. Unlike all of the adversarial training strategies,
our approach does not require computing adversarial images, and does thus not depend on a specific attack scheme. Instead, it aims to ensure a maximal separation between the different classes in high attention regions. This significantly differs from~\cite{mustafa2019adversarial}, which only clusters the penultimate layer's global representation, without focusing on discriminative regions and without attempting to separate these features.
Furthermore and more importantly, in contrast to all the above-mentioned methods, our approach is tailored to fine-grained recognition, making use of the representations that have proven effective in this field, such as Bags of Words (BoW)~\cite{jegou2010aggregating,wang2016learnable} and VLAD~\cite{arandjelovic2016netvlad,girdhar2017actionvlad}, which have the advantage over second-order features~\cite{yu2018statistically,kong2017low,gao2016compact} of providing some degree of interpretability.

\noindent{\textbf{Interpretability.}
Understanding the decisions of a DNN is highly important in real-world applications to build user trust. In the context of general image recognition, the trend of interpreting a DNN's decision was initiated by~\cite{zeiler2014visualizing}, followed by the popular CAMs~\cite{zhou2016learning}. Subsequently, variants of CAMs~\cite{selvaraju2017grad,chattopadhay2018grad} and other visualization strategies~\cite{nguyen2016synthesizing} were proposed.

Here, in contrast to these works, we focus on the task of fine-grained recognition. In this domain, BoW-inspired representations, such as the one of~\cite{nakka2018deep} and the ProtoPNet of~\cite{chen2019looks}, were shown to provide some degree of interpretability. However, the VLAD prototypes extracted by the method of~\cite{nakka2018deep} are not enforced to be class-specific. As such, while this method allows one to highlight the  image regions important for classification, it does not provide one with visual explanations of the network's decisions. 

 This is addressed by ProtoPNet~\cite{chen2019looks}, which extracts class-specific prototypes. However, the feature embedding learnt by ProtoPNet gives equal importance to all image regions, resulting in a large number of prototypes representing non-discriminative background regions, as illustrated by Fig.~\ref{fig:analysis}. Here, we overcome this by designing an attention-aware system that learns prototypes which are close to high-attention regions in feature space, while constraining the non-discriminative regions from all classes to be close to each other.  Furthermore, we show that this brings about not only interpretability, but also robustness to adversarial attacks, which has never been studied in the context of fine-grained recognition.

\cutsectionup
\section{Interpreting Adversarial Attacks }
\cutsectiondown
\label{sec:interpretation}
Before delving into our method, let us study in more detail the experiment depicted by Fig.~\ref{fig:analysis} to understand the decision of a fine-grained recognition CNN under adversarial attack. 
For this analysis, we experiment with two networks: the second-order attentional pooling network of~\cite{girdhar2017attentional} and the ProtoPNet of~\cite{chen2019looks}, both of which inherently encode some notion of interpretability  in their architecture, thus not requiring any post-processing.
Specifically,~\cite{girdhar2017attentional} uses class attention maps to compute class probabilities, whereas~\cite{chen2019looks} exploits the similarity between image regions and class-specific prototypes.
We analyze the reasons for the success of adversarial attacks on four images from the \emph{Black-footed albatross} class. 

As shown in Fig~\ref{fig:analysis}(a), under attack,~\cite{girdhar2017attentional} misclassifies all four images to \emph{Layman albatross}. Note that these two classes belong to the same general \emph{Albatross} family, and, for clean samples, the region with highest attention for these two classes is the bird's beak. Because the discriminative regions for these two classes correspond to the same beak region, which looks similar in both classes, the attack becomes easier as minimal perturbation is needed to change the class label. 

In the case of ProtoPNet~\cite{chen2019looks}, while the network also consistently misclassifies the attacked images, the resulting label differs across the different images, as shown in Fig.~\ref{fig:analysis} (b). In the top row, the situation is similar to that occurring with the method of~\cite{girdhar2017attentional}. By contrast, in the second row, the region activated in the input image corresponds to a different semantic part (wing) than that activated in the prototype (beak). Finally, in the last two rows, the network activates a background prototype that is common across the other categories and thus more vulnerable to attacks. 

In essence, the mistakes observed in Fig~\ref{fig:analysis} come from either the discriminative regions of two different classes being two close in feature space, or the use of non-discriminative regions for classification. This motivates us to encourage the feature representation of discriminative regions from different classes to be maximally separated from each other, while minimizing the influence of background regions by making use of attention and by encouraging the features in these regions to lie close to each other so as \emph{not} to be discriminative. This will complicate the attacker's task, by preventing their ability to leverage non-discriminative regions and forcing them to make larger changes in feature space to affect the prediction.

\cutsectionup
\section{Method}
\cutsectiondown

\providecommand{\localwidth}{}
\renewcommand{\localwidth}{\linewidth}

\providecommand{\localheight}{}
\renewcommand{\localheight}{8cm}

\begin{figure}[t]
	\footnotesize
	\centering
      \includegraphics[width=\textwidth]{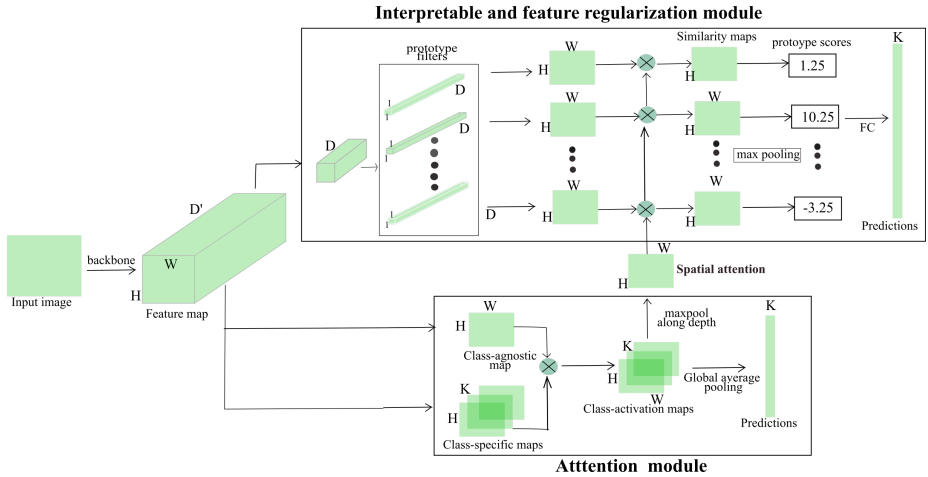}
	\caption{{\bf Overview of our framework}. Our approach consists of two modules acting on the features extracted by a backbone network. The attention module extracts attention maps that help the network to focus on the discriminative image regions. The feature regularization module further uses the attention maps to encourage separating the learned prototypes belonging to different classes.}
	\label{fig:architecture}
\end{figure}

In this section, we introduce our approach to increasing the robustness of fine-grained recognition by maximal separation of class-specific discriminative regions.  Figure~\ref{fig:architecture} gives an overview  our framework, which consists of two modules post feature extraction:
(i) An attention module that learns class-specific filters focusing on the discriminative regions; and (ii) a feature regularization module that maximally separates the class-specific features deemed discriminative by the attention module. 
Through the feature regularization module, we achieve the dual objective of providing interpretability and increasing the robustness of the backbone network to adversarial attacks.  

Note that, at inference time, we can either use the entire framework for prediction, or treat the attention module, together with the backbone feature extractor, as an standalone network. As will be demonstrated by our experiments, both strategies yield robustness to adversarial attacks, which evidences that our approach in fact robustifies the final feature map. Below, we first describe the overall architecture of our framework and then discuss the feature regularization module in more detail.

 \comment{
 \KN{ Few exps pending to support next sentence}
 Note that, the backbone network obtained by our formulation can be subsequently passed through a  general pooling layer such as fully connected layer or second order pooling and further finetuning the last layer.  Hence, our method can be thought as an general framework to achieve the robustness without any modification in network architecture. 
}

\cutsectionup
\subsection{Architecture}
\cutsubsectiondown

Formally, let $\bI_i$ denote an input image, and $\bX_i \in \mathbb{R}^{H \times W \times D'}$ represent the corresponding feature map extracted by a fully-convolutional backbone network. Our architecture is inspired by the ProtoPNet of~\cite{chen2019looks}, in the sense that we also rely on class-specific prototypes. However, as shown in Section~\ref{sec:interpretation}, ProtoPNet fails to learn discriminative prototypes, because it allows the prototypes to encode non-discriminative background information and to be close in feature space even if they belong to different classes. To address this, we propose to focus on the important regions via an attention mechanism and to regularize the prototypes during training.

Specifically, our attention module consists of two sets of filters: (i) A class-agnostic $1\times 1 \times D'$ filter yielding a single-channel map of size $H \times W$;  and (ii) $K$ class-specific $1 \times 1 \times D'$ filters producing $K$ maps of size $H \times W$ corresponding to the $K$ classes in the dataset. Each of the class-specific map is then multiplied by the class-agnostic one, and the result is spatially averaged to generate a $K$-dimensional output. As shown in~\cite{girdhar2017attentional}, this multiplication of two attention maps is equivalent to a rank-1 approximation of second-order pooling, which has proven to be well-suited to aggregate local features for fine-grained recognition.  

The second branch of our network extracts interpretable prototypes and is responsible to increase the robustness of the features extracted by the backbone. To this end, $\bX_i$ is first processed by two $1 \times 1$ convolutional layers to decrease the channel dimension to $D$. The resulting representation is then passed through a prototype layer that contains $m$ learnable prototypes  of size $1 \times 1 \times D$, resulting in $m$ similarity maps of size $H \times W$. Specifically, the prototype layer computes the $L^2$ distance between each local feature and each prototype, and passes this distance through an activation function defined as  $f(r) = \log\left((\|r\|_2^2 + 1)/(\|r\|_2^2 + \gamma)\right)$, where $\gamma$ is set to $1e-5$.
In contrast to~\cite{chen2019looks}, to focus on discriminative regions, we modulate the resulting similarity maps with an attention map $\bA_i$, computed by max-pooling the final class-specific maps of the attention module. We then spatially max-pool the resulting attention-aware similarity maps to obtain similarity scores, which are passed through the final classification layer to yield class probabilities. As in~\cite{chen2019looks}, we make the prototypes class specific by splitting the $m$ prototypes into $K$ sets of $c$ prototypes and initializing the weights of the classification layer of the prototype branch to +1 for positive connections between prototype and class label and -0.5  for negative ones. 
While exploiting attention encourages the prototypes to focus on the discriminative regions, nothing explicitly prevents prototypes from different classes to remain close in feature space, thus yielding a small margin between different classes and making the classifier vulnerable to attacks. This is what we address below.


\cutsubsectionup
\subsection{Discriminative Feature Separation}
\cutsubsectiondown
To learn a robust feature representation, we introduce two feature regularization losses that aim to maximally separate the prototypes of different classes. Let $\bx_i^t$ represent a local feature vector at location $t$ in feature map $\bX_i$ from image $\bI_i$ with label $y_i$.  Furthermore, let $N = W \cdot H$ be the total number of feature vectors in $\bX_i$, and $\mathbf{P}_{y_i}$ be the set of prototypes belonging to class $y_i$.

Our regularization consists of two attention-aware losses, a clustering one and a separation one. The attentional-clustering loss pulls the high-attention regions in a sample close to the nearest prototype of its own class.  We express this as
\begin{footnotesize}
	\begin{equation}
	\begin{split}
	\textrm{L}_{clst}^{att}(\bI_i) = \sum_{t=1}^{N} a_{i}^t  \min_{l: \mathbf{p}_l \in \mathbf{P}_{y_i}}  \| \mathbf{x}_i^t - \mathbf{p}_l\|_2^2\;,
	\end{split}
	\label{eq:loss1}
	\end{equation}
\end{footnotesize}
where $a_{i}^t$ is the attention value at location $t$ in $\bA_i$.
By contrast, the attentional-separation loss pushes the high-attention regions away from the nearest prototype of any other class. We compute it as
\begin{footnotesize}
\begin{equation}
\textrm{L}_{sep}^{att}(\bI_i) = -\sum_{t=1}^{N} a_{i}^t  \min_{l: \mathbf{p}_l \not\in \mathbf{P}_{y_i}}  \|\mathbf{x}_i^t - \mathbf{p}_l\|_2^2\;.
\label{eq:loss2}
\end{equation}
\end{footnotesize}

While these two loss functions encourage the prototypes to focus on high-attention, discriminative regions, they leave the low-attention regions free to be close to any prototype, thus increasing the vulnerability of the network to attacks. We therefore further need to push the non-discriminative regions away from such informative prototypes. A seemingly natural way to achieve this would consist of exploiting inverted attention maps, such as $1-\ba_{t}$ or $1/\ba_{t}$. However, in practice, we observed this to make training unstable. Instead, we therefore propose to make use of the attention maps from \emph{other} samples to compute the loss for sample $i$.
Specifically, we re-write our regularization loss for sample $i$ as
\begin{footnotesize}
	\begin{equation}
	\textrm{L}_{reg}(\bI_i) =  \sum_{j=1}^{B}  \sum_{t=1}^{N}  \lambda_{1} a_{j}^t  \min_{l: \mathbf{p}_l \in \mathbf{P}_{y_i}}  \| \mathbf{x}_i^t - \mathbf{p}_l\|_2^2   - \lambda_{2}  a_{j}^t  \min_{l: \mathbf{p}_l \not\in \mathbf{P}_{y_i}}  \|\mathbf{x}_i^t - \mathbf{p}_l\|_2^2 \;,
	\label{eq:loss3}
	\end{equation}
\end{footnotesize}
where $B$ is the number of samples in the mini-batch. When $j=i$, we recover the two loss terms defined in Eqs.~\ref{eq:loss1} and~\ref{eq:loss2}. By contrast, when $j\neq i$, we exploit the attention map of a different sample.
The intuition behind this is that either the attention map of sample $j$ focuses on the same regions as that of sample $i$, and thus the loss serves the same purpose as when using the attention of sample $i$, or it focuses on other regions, and the loss then pushes the corresponding feature map, encoding a low-attention region according to the attention map of sample $i$, to its own prototype in class $y_i$. In practice, we have observed this procedure to typically yield a single background prototype per class. These background prototypes inherently become irrelevant for classification because they correspond to low-attention regions and have thus a similarity score close to zero, thanks to our attention-modulated similarity maps. As such, we have empirically found that all background prototypes tend to cluster.

Ultimately, we write our total loss function for sample $i$ as
\begin{footnotesize}
	\begin{equation*}
	\textrm{L}(\bI_i) =  \textrm{CE}_{att}(\bI_i) + \textrm{CE}_{reg}(\bI_i) + \textrm{L}_{reg}(\bI_i)\;,
	\label{eq:loss4}
	\end{equation*}
\end{footnotesize}
where $\textrm{CE}_{att}$ and $\textrm{CE}_{reg}$ represent the cross-entropy loss of the attention module and the feature regularization module, respectively.

At inference time, we perform adversarial attacks on the joint system by exploiting the cross-entropy loss of both the attention and feature regularization module. Furthermore, we also attack the attention module on its own, showing that, together with the feature extraction backbone, it can be used as a standalone network and also inherits robustness from our training strategy.


\cutsectionup
\section{Experiments}
\cutsectiondown
In this section, we  evaluate our approach on several fine-grained recognition benchmark datasets.  Below, we first discuss the details of our experimental setting, and then evidence the benefits of our approach over several baselines both quantitatively and qualitatively.

\cutsubsectionup
\subsection{Experimental Setting}
\cutsubsectiondown
\noindent{\emph{Datasets.}} We experiment on two popular fine-grained datasets, Caltech UCSD Birds (CUB)~\cite{cub200} and Stanford Cars-196~\cite{cars196}.  CUB consists of 11,788 images from 200 fine-grained categories of birds with large variations in pose, split into  5394 training and 5794 testing images. Stanford Cars-196 contains 16,185 images from 196 classes, with 8144 training and 8041 testing images. For all our experiments, we use the standard train-test splits released with the datasets. Similarly to~\cite{chen2019looks}, we perform data augmentation offline on every image in the dataset, using random rotation, distortion, and skew and flip operations to obtain a 30 fold increase in dataset size. For evaluation, we report the accuracy over the entire dataset as in~\cite{chen2019looks,girdhar2017attentional}. 

\vspace{0.1cm}
\noindent{\emph{Threat Model.}}  We consider both white-box and black-box attacks under an $\ell_{\infty}$-norm budget. We evaluate robustness for two attack tolerances $\epsilon = \{2/255, 8/255\}$. In addition to the popular 10-step PGD attack~\cite{madry2017towards}, we test our framework with FGSM~\cite{fgsm}, BIM~\cite{bim}, and MI~\cite{dong2018boosting} attacks. For PGD attacks, we set the step size  $\alpha$ to $1/255$ for $\epsilon = 2/255$  and to $2/255$ for  $\epsilon= 8/255$. For the other attacks, we set number of iterations to 10 and the step size $\alpha$ to $\epsilon$ divided by the number of iterations, as in~\cite{mustafa2019adversarial,mao2019metric}.   For black-box attacks, we transfer the adversarial examples generated using 10-step PGD with $\epsilon=8/255$ and $\alpha=2/255$  on either a similar VGG16~\cite{vgg16} architecture, or a completely different DenseNet-121~\cite{densenet} architecture. We denote by BB-V and BB-D the black-box  attacks transferred from VGG16 and DenseNet-121, respectively.

\vspace{0.1cm}
\noindent{\emph{Networks.} We evaluate our approach using 3 backbone networks: VGG-16~\cite{vgg16}, VGG-19~\cite{vgg16} and ResNet-34~\cite{resnet}. Similarly to~\cite{chen2019looks}, we perform all experiments on images cropped according to the bounding boxes provided with the dataset, and resize the resulting images to $224 \times 224$. For both VGG-16 and VGG-19, we use the convolutional layers until the 4th block to output $7\times7$ spatial maps of 512 channels. For ResNet-34, we take the network excluding the final global average pooling layer as backbone. We initialize the backbone networks with weights pretrained on ImageNet~\cite{krizhevsky2012imagenet}.

\vspace{0.1cm}
\noindent{\emph{Evaluated Methods.}}
As baselines, we use the attentional pooling network {\bf (AP)} of~\cite{girdhar2017attentional},  and the state-of-the-art {\bf ProtoPNet} of~\cite{chen2019looks}. We use \textbf{Ours-FR} and \textbf{Ours-A} to denote the output of our feature regularization module and of our attention module, respectively. In other words, AP and Ours-A share the same architecture at inference time, and Ours-FR is an attention-aware variant of ProtoPNet.

To further boost the performance of the baselines and of our approach, we perform adversarial training. Specifically, we generate adversarial examples using the recent fast adversarial training strategy of~\cite{wong2020fast}, which relies on a single step FGSM with random initialization. During training, we set $\epsilon$ to $8/255$ and $\alpha$ to $1.25\epsilon$ as suggested in~\cite{wong2020fast}. This was shown in~\cite{wong2020fast} to perform on par with PGD-based adversarial training, while being computationally much less expensive. For our approach, during fast adversarial training, we use the cross-entropy loss of both modules to generate the adversarial images. We denote by {\bf AP-AT} and {\bf ProtoPNet-AT} the adversarially-trained AP and ProtoPNet baselines, respectively, and by \textbf{Ours-FR+AT} and \textbf{Ours-A+AT} the adversarially-trained counterparts of our two sub-networks.
	
\vspace{0.1cm}
\noindent{\emph{Training Details.}} We implemented our approach using the PyTorch library, and ran our experiments on a single 32GB Tesla GPU. We set the mini-batch size $B$ to $75$ during training. We use $c=10$ prototypes per class, resulting in a total of  $m=2000$ and $m=1960$ prototypes for  CUB~\cite{cub200} and Cars~\cite{cars196}, respectively. 
We set $\lambda_{1}$ to $\{10,100\}$ and $\lambda_{2}=0.08$ depending on the dataset and architecture. We first fine-tune the attention and feature regularization modules, except for the classification layer of the latter, for 5 epochs with a learning rate of $0.0003$, keeping the backbone network fixed. We then jointly train all the layers, except the feature regularization classifier,  to minimize the objective of Eq.~\ref{eq:loss3} for 25 epochs, with an initial learning rate of 0.003 and a decay rate of 0.1 applied every 10 epochs. After 30 epochs, we project the prototypes to the nearest training image patch of the same class and optimize the classification layer of the feature regularization module for 15 epochs. We use Adam~\cite{kingma2014adam} with the default momentum values for all our experiments. 

\cutsubsectionup
\subsection{Results on CUB 200}
\cutsectiondown
\noindent{\emph{Quantitative Analysis.}}
We first compare the accuracy of our method to that of the baselines with the three backbone networks on CUB200. Table~\ref{tab:birdsclean} and Table~\ref{tab:birdsAT} provide the results for vanilla and fast adversarial training, respectively.  On the clean samples, \textbf{Our-A} yields the best accuracy across all backbones, and \textbf{Ours-FR} typically surpasses its non-attentional counterpart {\bf ProtoPNet}~\cite{chen2019looks}.  This is true both without (Table~\ref{tab:birdsclean}) and with (Table~\ref{tab:birdsAT}) adversarial training.

Under adversarial attack, our approach, without and with adversarial training, yields better robustness under almost all attacks and backbones. Importantly, the boost in performance is larger for attacks with larger perturbations. Furthermore, our model trained with clean samples sometimes outperform even the adversarially-trained baselines. For example, on VGG-16 with PGD attack with $\epsilon=8/255$, {\bf AP+AT} yields an accuracy of $16.9\%$  (Table~\ref{tab:birdsAT}) while \textbf{Ours-A} reaches $21.8\%$ accuracy (Table~\ref{tab:birdsclean}). This evidences the ability of our feature regularization module to learn robust features, even without seeing any adversarial examples. This is further supported by the fact that, despite {\bf AP} and \textbf{Ours-A} having the same architecture at inference, \textbf{Ours-A} is more robust to attacks. {Furthermore, in contrast to adversarially-trained models, our vanilla approach does not trade off clean accuracy for robustness. For example, on VGG-16, while adversarial training made the clean accuracy of {\bf AP+AT} drop to 54.9\% (Table~\ref{tab:birdsAT}), that of \textbf{Ours-A} is 80.4\% (Table~\ref{tab:birdsclean}). 
In other words, we achieve good robustness \emph{and} clean accuracy.}

\begin{table}[t]
\scriptsize
  \centering
 \setlength\tabcolsep{1.4pt}
  \begin{tabular}{llllllllllllllll}
    \midrule
      Attacks  & Clean & FGSM& FGSM &  BIM  & BIM  & PGD  &  PGD  &  MIM & MIM  & BB-V & BB-D \\    
(Steps,$\epsilon$) &(0,0)  & (1,2) & (1,8) & (10,2)& (10,8) & (10,2) & (10,8)& (10,2) & (10,8)  & (10,2) & (10,8) \\ 
   
     \midrule
    \multicolumn{10}{c}{\textbf{VGG16}} \\
    \midrule

   {AP}~\cite{girdhar2017attentional} &78.0\%&36.5\%&31.0\%&27.7\%&14.6\%&23.5\%&11.7\%&30.2\%& 16.7\%& 9.6\% & 60.4\%\\
   {ProtoPNet}~\cite{chen2019looks}&69.0\%&19.9\%&8.10\%&3.80\%&0.00\%&2.20\%&0.00\%&5.0\%&0.10\% & \textbf{22.9\%} & 58.5\%\\
   \textbf{Ours-A}&\textbf{80.4\%}&47.2\%&40.2\%&40.0\%&23.2\%&35.3\%&21.8\%&42.2\%&26.4\%& 12.9\% & \textbf{66.9\%}\\
   \textbf{Ours-FR}&73.2\%&\textbf{49.9\%}&\textbf{42.2\%}&\textbf{42.5\%}&\textbf{35.3\%}&\textbf{38.4\%}&\textbf{30.1\%}&\textbf{42.9\%}&\textbf{37.5\%} & 15.4\% & {59.7\%}\\
    \midrule
   \multicolumn{10}{c}{\textbf{VGG19}} \\
   \midrule
  {AP}~\cite{girdhar2017attentional}& 75.7\%& 20.4\%&14.5\%&13.4\%&6.9\%&10.5\%&5.7\%&14.8\%&6.9\% & 21.1\% & 61.3\%\\
 {ProtoPNet}~\cite{chen2019looks}&73.8\%&22.9\%&11.1\%&3.2\%& 0.0\%& 1.2\% &0.0\% &3.6\%&0.0\% & 21.0\% & 58.0\%\\
 \textbf{Ours-A}&\textbf{79.7\%}&51.4\%&44.6\%&42.3\%&26.5\%&36.8\%&26.3\%& \textbf{45.0\%}&\textbf{42.6\%}& 29.8\% & \textbf{68.2\%}\\
 \textbf{Ours-FR}&75.4\%&\textbf{52.2\%}&\textbf{46.3\%}&\textbf{46.6\%}&\textbf{41.3\%}&\textbf{42.4\%}&\textbf{31.0\%}&{44.4\%}&{37.6\%}& \textbf{30.4\%} & 63.7\%\\
    \midrule
   \multicolumn{10}{c}{\textbf{ResNet-34}} \\
   \midrule
   
     {AP}~\cite{girdhar2017attentional} &79.9\%&30.4\%&\textbf{26.3\%}&18.0\%&7.20\%&13.2\%&5.8\%&22.3\%&8.6\% & 43.0\%& 59.4\%\\
   {ProtoPNet}~\cite{chen2019looks}&75.1\% &23.2\%&12.8\%&7.80\%&1.80\%&4.10\%&1.00\%&8.90\%&2.20\%& 39.1\%& 53.0\%\\
   \textbf{Ours-A}&\textbf{80.3\%}&\textbf{31.6\%}&20.1\%&17.4\%&9.90\%&13.5\%&9.20\%&19.9\% & 11.4\% & \textbf{47.9\%}& \textbf{64.6\%}\\
   \textbf{Ours-FR}&75.2\%&31.2\%&{23.9\%}&\textbf{22.2\%}&\textbf{20.2\%}&\textbf{21.4\%}&\textbf{18.2\%}&\textbf{23.4\%}&\textbf{20.8\%} &43.7\%& 58.9\%\\

   \bottomrule
  \end{tabular}
  \vspace{1mm}
  \caption{\small{Classification accuracy of different networks with $\ell_{\infty}$ based  attacks on CUB200. The best result of each column and each backbone is shown in bold.\MS{Double-check the bold numbers. One was wrong.} The last two columns correspond to black-box attacks.}}
  \label{tab:birdsclean}
\end{table}

\begin{table}[t]
	\scriptsize
	\centering
	\setlength\tabcolsep{0.5pt}
	\begin{tabular}{llllllllllll}
		\midrule
		Attacks  & Clean & FGSM& FGSM &  BIM  & BIM  & PGD  &  PGD  &  MIM & MIM & BB-V & BB-D\\    
		(Steps,$\epsilon$) & (0,0)  & (1,2) & (1,8) & (10,2)& (10,8) & (10,2) & (10,8)& (10,2) & (10,8) & (10,2) & (10,8)  \\ 
		
		\midrule
		\multicolumn{10}{c}{\textbf{VGG16}} \\
		\midrule
		
		{AP+AT}~\cite{girdhar2017attentional} & 54.9\% & 44.9\%&24.2\% &41.9\% & 18.2\% &41.2\% &16.9\%& 41.9\% &18.7\% & 54.6\% & 54.0\%\\		{ProtoPNet+AT}~\cite{chen2019looks}& 60.1\%& 44.5\% & 26.9\% &\textbf{57.1\%} &10.9\% & 35.9\% &10.3\% &37.6\% &13.5\% & 58.4\%& 59.1\%\\
		\textbf{Ours-AP+AT} & \textbf{63.1\%} &\textbf{56.1\%}& 34.8\%& {51.7\%}& 29.6\%&\textbf{50.8\%}	& 28.0\%	& \textbf{52.0\%}&\textbf{32.5\%} & \textbf{66.3\%} & \textbf{68.0\%} \\
 \textbf{Ours-P+AT}& 63.0\% &53.3\% &\textbf{37.3\%}& 49.4\% &\textbf{30.4\%} &48.1\% &\textbf{28.6\%}& 49.7\% &31.1\% & 61.1\% & 62.0\%\\

			\midrule
		\multicolumn{10}{c}{\textbf{VGG19}} \\
		\midrule
		
		{AP+AT}~\cite{girdhar2017attentional}& 0.0\% & 0.0\% &0.0\% &0.0\% &0.0\% &0.0\% &0.0\% &0.0\% &0.0\% & 0\% & 0\%  \\
		{ProtoPNet+AT}~\cite{chen2019looks}& 55.1\%& 40.0\%&28.9\% & 26.5\% &11.3\% & 29.7\% &9.60\% &25.6\% &10.2\% & 53.6\%& 53.9\%\\
		\textbf{Ours-A+AT} & \textbf{68.2\%} &\textbf{57.1\%}& {36.5\%}& \textbf{53.2\%}& 30.4\%&\textbf{52.6\%}	& \textbf{29.2\%}	& \textbf{53.5\%}&31.2\% & \textbf{66.2\%} & \textbf{66.9\%}\\
		\textbf{Ours-FR+AT}& 64.4\% &55.5\% &\textbf{37.4\%}& 51.2\% &\textbf{30.6\%} &50.4\% &28.7\%& 52.1\% & \textbf{32.3\%} & 62.5\% &63.2\%\\
		
			\midrule
		\multicolumn{10}{c}{\textbf{ResNet-34}} \\
		\midrule
		
		{AP+AT}~\cite{girdhar2017attentional} & 55.6\% & 47.8\% &29.2\% &44.80\% &21.0\% &44.5\% &19.4\% &44.9\% &21.9\%  & 55.30\%& 55.2\%\\
		{ProtoPNet+AT}~\cite{chen2019looks}& 57.9\%& 46.5\%&30.3\% & 41.1\% &21.1\% & 40.3\% &18.4\% &41.5\% &20.9\% & 56.9\%& 57.0\%\\
		\textbf{Ours-A+AT} & \textbf{62.2\%} &\textbf{54.2\%}& \textbf{35.7\%}& \textbf{51.5\%}& \textbf{25.5\%}&\textbf{51.0\%}	& \textbf{23.1\%}	& \textbf{51.6\%}&\textbf{26.6\%} & \textbf{61.5\%} & \textbf{61.9\%}\\
		\textbf{Ours-FR+AT}& 57.6\% &49.5\% &32.3\%& 45.8\% &23.2\% &44.9\% &19.9\%& 46.1\% &24.6\% & 57.1\% &57.0\%\\
		\bottomrule
	\end{tabular}
	\vspace{1mm}
\caption{{\small Classification accuracy of different robust networks with $\ell_{\infty}$ based  attacks on CUB200. The best result of each column and each backbone is shown in bold. The last two columns correspond to black-box attacks. Note that AP+AT on VGG19 did not converge.} }
	\label{tab:birdsAT}
\end{table}

\vspace{0.1cm}
\noindent{\emph{Transferability Analysis.}}
To evaluate robustness to black-box attacks, we transfer adversarial examples generated from substitute networks to our framework and to the baselines. As substitute models, we use a VGG-16~\cite{vgg16} and DenseNet-121~\cite{densenet} backbone followed by global average pooling and a classification layer.  The corresponding results are reported in the last two columns of Table~\ref{tab:birdsclean} and Table~\ref{tab:birdsAT}. As in the white-box case, our approach outperforms the baselines in this black-box setting, thus confirming its effectiveness at learning robust features.

\vspace{0.1cm}
\noindent{\emph{Qualitative Analysis.}}
Let us now qualitatively evidence the benefits of our approach. To this end, in Figure~\ref{fig:protobirds}, we visualize the 10 class-specific prototypes learned by ProtPNet and by our approach for the  \emph{Blackfooted albatross} class. Specifically, we show the activation heatmaps of these prototypes on the source image that they have been projected to. Note that ProtoPNet learns multiple background prototypes, whereas our approach encodes all the background information in a single non-discriminative prototype. Furthermore, ProtoPNet~\cite{chen2019looks} focuses on much larger regions, which can be expected to be less discriminative than the fine-grained regions obtained using our approach. This is due to our use of attention, which helps the prototypes to focus on the areas that are important for classification.


\providecommand{\localwidth}{}
\renewcommand{\localwidth}{\linewidth}

\providecommand{\localheight}{}
\renewcommand{\localheight}{8cm}

\begin{figure}[b]
	\footnotesize
	\centering
      \includegraphics[width=\textwidth]{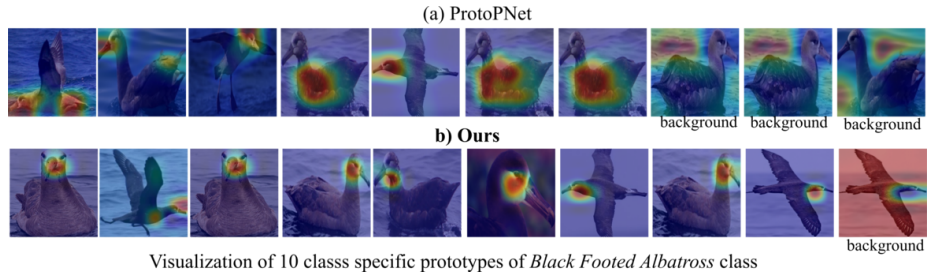}
	\caption{{\bf Comparison of the prototypes learned with ProtoPNet~\cite{chen2019looks} and with our approach on CUB}. ProtoPNet yields multiple background prototypes and prototypes that focus on large regions. By contrast, our prototypes are finer-grained and thus more representative of the specific class in the images.}
	\label{fig:protobirds}
\end{figure}


\providecommand{\localwidth}{}
\renewcommand{\localwidth}{\linewidth}

\providecommand{\localheight}{}
\renewcommand{\localheight}{7cm}

\begin{figure}[t]
	\footnotesize
	\centering		
		\includegraphics[height= \localheight,width=\localwidth]{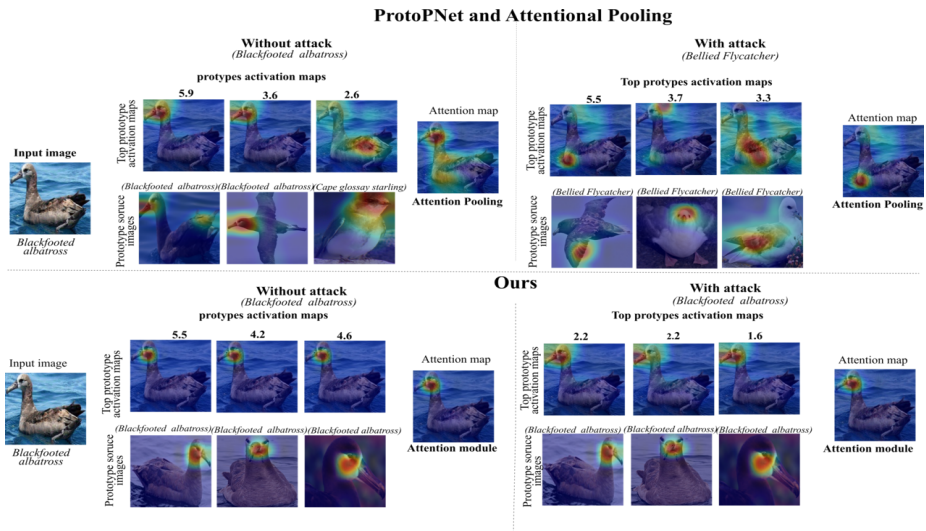}
	\caption{{\bf Comparison of the activated image regions without and with attack.} Without attack, the baselines (AP and ProtoPNet) tend to rely on relatively large regions, sometimes corresponding to wrong classes, for prediction. By contrast, our approach focuses more closely on the discriminative regions. Under attack, this phenomenon is further increased, with ProtoPNet and AP activating incorrect prototypes and regions. The activations obtained with our approach remain similar to those obtained without attacks, albeit with a decrease in the similarity scores, indicated above the top prototype activation maps.
}\label{fig:exp1}
\end{figure}

In Figure~\ref{fig:exp1}, we analyze the effect of adversarial attacks on AP, ProtoPNet and our approach (all without adversarial training) by visualizing the attention maps and/or  a few top activated prototypes along with their similarity scores for a \emph{Blackfooted albatross} image with and without attack.  
Without attack, AP activates a larger region than our attention module. Furthermore, ProtoPNet activates a prototype from a different class (\emph{Cape glossay starling}), while our approach focuses on the correct class only.
This already shows that the features learned by these baselines are less discriminative, making them more vulnerable to adversarial attacks. As a matter of fact, under attack, AP focuses on a different region that is not discriminative for the \emph{Blackfooted albatross} class. Similarly, ProtoPNet activates prototypes of different classes with high similarity scores, highlighting non-discriminative regions.  By contrast, the prototypes activated by our approach remain the same as in the clean case, thus corresponding to the correct class. Nevertheless, the similarity scores drop significantly. 
{This suggests a potential strategy to detect an attack by analyzing the magnitude of the similarity scores, which we leave for future work.}

\cutsubsectionup
\subsection{Results on Stanford Cars}
\cutsubsectiondown
We now present on results on Stanford Cars~\cite{cars196}. In Table~\ref{tab:carsclean}, we report the results obtained using vanilla training. As in the CUB case, our approach yields better robustness than the baselines. We provide a qualitative analysis and the results obtained with adversarial training in the supplementary material.

\providecommand{\localwidth}{}
\renewcommand{\localwidth}{\linewidth}

\providecommand{\localheight}{}
\renewcommand{\localheight}{8cm}

\begin{figure}[t]
	\footnotesize
	\centering
      \includegraphics[width=\textwidth]{\Introfig{proto2_resize.png}}
	\caption{{\bf Comparison of the prototypes learned with ProtoPNet~\cite{chen2019looks} and with our approach on Stanford-Cars}. ProtoPNet yields prototypes that cover large regions, whereas our prototypes more focused.}
	\label{fig:proto_cars}
\end{figure}

In Figure~\ref{fig:proto_cars}, we compare the prototypes  learned with ProtoPNet and with our approach for the \emph{Accura TL Sedan} class. As before, while the prototypes learned by ProtoPNet cover large regions, those obtained with our framework are more focused on the discriminative parts of the car.

\begin{table}[t]
\scriptsize
  \centering
 \setlength\tabcolsep{1pt}
  \begin{tabular}{llllllllllllllll}
    \midrule
      Attacks  & Clean & FGSM& FGSM &  BIM  & BIM  & PGD  &  PGD  &  MIM & MIM &  BB-V & BB-D\\    
(Steps,$\epsilon$) &(0,0)   & (1,2) & (1,8) & (10,2)& (10,8) & (10,2) & (10,8)& (10,2) & (10,8)& (10,2) & (10,8)  \\ 
   
     \midrule
    \multicolumn{11}{c}{\textbf{VGG16}} \\
    \midrule
    
    {AP}~\cite{girdhar2017attentional} & \textbf{91.2\%}&52.6\%&40.2\%&37.4\%&10.5\%&28.8\%&6.93\%&41.7\%&12.9\% & 12.5\% & \textbf{82.5\%} \\
    {ProtoPNet}~\cite{chen2019looks}&84.5\%&31.2\%&9.85\%&4.78\%&0.01\%&2.23\%&0.00\%&6.5\%&0.01\%& \textbf{27.8\%} & 75.5\%\\
   \textbf{Ours-A}&{88.5\%}&58.7\%&40.2\%&48.0\%&28.6\%&46.5\%&21.7\%&\textbf{53.2\%}&33.2\%& 19.9\% & {82.2\%}\\
   \textbf{Ours-FR}&83.8\%&\textbf{60.1\%}&\textbf{52.0\%}&\textbf{51.3\%}&\textbf{41.0\%}&\textbf{47.8\%}&\textbf{32.9\%}&{51.8\%}&\textbf{43.9\%}& {23.4\%} & 75.1\%\\
   \midrule
   \multicolumn{11}{c}{\textbf{VGG19}} \\
   \midrule
   {AP} & \textbf{91.5\%}& 50.1\%&37.8\%&33.4\%&10.3\%&23.83\%&6.93\%&37.9\%&12.7\%& 20.7\% & \textbf{82.8\%}\\
   {ProtoPNet}~\cite{chen2019looks}&85.6\%&34.1\%&20.8\%&11.3\%&1.11\%& 4.40\%& 0.5\% &14.2\% &1.39\% & 26.5\% & 75.5\%\\
   \textbf{Ours-A}&{88.7\%}&\textbf{64.4\%}&\textbf{54.8\%}&\textbf{56.4\%}&36.7\%&\textbf{51.7\%}&33.4\%&\textbf{58.1\%}&41.0\%& 35.9\% & 82.5\%\\
   \textbf{Ours-FR}&85.0\%&{62.4\%}&{54.7\%}&{54.5\%}&\textbf{45.7\%}&{51.2\%}&\textbf{38.5\%}&{54.3\%}&\textbf{47.6\%} & \textbf{36.1\%} & 76.8\%\\
   \midrule
   \multicolumn{11}{c}{\textbf{ResNet-34}} \\
   \midrule
   
   {AP} & \textbf{90.5\%}&{38.8\%}&31.7\%&23.2\%&8.12\%&16.3\%&6.6\%&28.5\%&10.1\% & 50.8\%&  76.6\%\\
   {ProtoPNet}~\cite{chen2019looks}&88.3\% &33.1\%&15.3\%&8.9\%&3.6\%&6.1\%&3.4\%&9.6\%&3.6\%& 45.0\% & 70.6\%\\
   \textbf{Ours-A}&{89.6\%}&{37.0\%}&29.2\%&25.4\%&16.2\%&20.6\%&14.8\%&27.8\% & 17.9\%& \textbf{52.7\%} & \textbf{79.1\%}\\
   \textbf{Ours-FR}&87.2\%&\textbf{40.6\%}&\textbf{32.0\%}&\textbf{29.1\%}&\textbf{24.2\%}&\textbf{26.5\%}&\textbf{18.8\%}&\textbf{29.8\%}&\textbf{24.9\%} & 48.4\% & 74.3\%\\

   \bottomrule
  \end{tabular}
  \vspace{1mm}
\caption{{\small Classification accuracy of different undefended networks with $\ell_{\infty}$ based  attacks on Cars196. The best result of each column and each backbone is shown in bold. The last two columns correspond to black-box attacks.}}
  \label{tab:carsclean}
 \vspace{-7mm}
\end{table}

\cutsectionup
\section{Conclusion}
\cutsectiondown
In this paper, we have performed the first study of adversarial attacks for fine-grained recognition. Our analysis has highlighted the key factor for the success of adversarial attacks in this context. This has inspired us to design an attention- and prototype-based framework that explicitly encourages the prototypes to focus on the discriminative image regions. Our experiments have evidenced the benefits of our approach, able to match and sometimes even outperform adversarial training, despite not requiring seeing adversarial examples during training. An interesting observation arising from our experiments is that, even when our method is robust to an attack, the similarity scores it computes tend to be lower. In the future, we will therefore investigate if this can be leveraged to design an attack detection mechanism.
	
	\bibliographystyle{splncs04}
	\bibliography{egbib}

\begin{thebibliography}{10}
\providecommand{\url}[1]{\texttt{#1}}
\providecommand{\urlprefix}{URL }
\providecommand{\doi}[1]{https://doi.org/#1}

\bibitem{arandjelovic2016netvlad}
Arandjelovic, R., Gronat, P., Torii, A., Pajdla, T., Sivic, J.: Netvlad: Cnn
  architecture for weakly supervised place recognition. In: Proceedings of the
  IEEE conference on computer vision and pattern recognition. pp. 5297--5307
  (2016)

\bibitem{athalye2018obfuscated}
Athalye, A., Carlini, N., Wagner, D.: Obfuscated gradients give a false sense
  of security: Circumventing defenses to adversarial examples. arXiv preprint
  arXiv:1802.00420  (2018)

\bibitem{carlini2017towards}
Carlini, N., Wagner, D.: Towards evaluating the robustness of neural networks.
  In: 2017 ieee symposium on security and privacy (sp). pp. 39--57. IEEE (2017)

\bibitem{chattopadhay2018grad}
Chattopadhay, A., Sarkar, A., Howlader, P., Balasubramanian, V.N.: Grad-cam++:
  Generalized gradient-based visual explanations for deep convolutional
  networks. In: 2018 IEEE Winter Conference on Applications of Computer Vision
  (WACV). pp. 839--847. IEEE (2018)

\bibitem{chen2019looks}
Chen, C., Li, O., Tao, D., Barnett, A., Rudin, C., Su, J.K.: This looks like
  that: deep learning for interpretable image recognition. In: Advances in
  Neural Information Processing Systems. pp. 8928--8939 (2019)

\bibitem{dong2018boosting}
Dong, Y., Liao, F., Pang, T., Su, H., Zhu, J., Hu, X., Li, J.: Boosting
  adversarial attacks with momentum. In: Proceedings of the IEEE conference on
  computer vision and pattern recognition. pp. 9185--9193 (2018)

\bibitem{gao2016compact}
Gao, Y., Beijbom, O., Zhang, N., Darrell, T.: Compact bilinear pooling. In:
  Proceedings of the IEEE conference on computer vision and pattern
  recognition. pp. 317--326 (2016)

\bibitem{girdhar2017attentional}
Girdhar, R., Ramanan, D.: Attentional pooling for action recognition. In:
  Advances in Neural Information Processing Systems. pp. 34--45 (2017)

\bibitem{girdhar2017actionvlad}
Girdhar, R., Ramanan, D., Gupta, A., Sivic, J., Russell, B.: Actionvlad:
  Learning spatio-temporal aggregation for action classification. In:
  Proceedings of the IEEE Conference on Computer Vision and Pattern
  Recognition. pp. 971--980 (2017)

\bibitem{fgsm}
Goodfellow, I.J., Shlens, J., Szegedy, C.: Explaining and harnessing
  adversarial examples. arXiv preprint arXiv:1412.6572  (2014)

\bibitem{resnet}
He, K., Zhang, X., Ren, S., Sun, J.: Deep residual learning for image
  recognition. In: Proceedings of the IEEE conference on computer vision and
  pattern recognition. pp. 770--778 (2016)

\bibitem{densenet}
Huang, G., Liu, Z., Van Der~Maaten, L., Weinberger, K.Q.: Densely connected
  convolutional networks. In: Proceedings of the IEEE conference on computer
  vision and pattern recognition. pp. 4700--4708 (2017)

\bibitem{jegou2010aggregating}
J{\'e}gou, H., Douze, M., Schmid, C., P{\'e}rez, P.: Aggregating local
  descriptors into a compact image representation. In: 2010 IEEE computer
  society conference on computer vision and pattern recognition. pp.
  3304--3311. IEEE (2010)

\bibitem{kannan2018adversarial}
Kannan, H., Kurakin, A., Goodfellow, I.: Adversarial logit pairing. arXiv
  preprint arXiv:1803.06373  (2018)

\bibitem{karpathy2014large}
Karpathy, A., Toderici, G., Shetty, S., Leung, T., Sukthankar, R., Fei-Fei, L.:
  Large-scale video classification with convolutional neural networks. In:
  Proceedings of the IEEE conference on Computer Vision and Pattern
  Recognition. pp. 1725--1732 (2014)

\bibitem{kingma2014adam}
Kingma, D.P., Ba, J.: Adam: A method for stochastic optimization. arXiv
  preprint arXiv:1412.6980  (2014)

\bibitem{kong2017low}
Kong, S., Fowlkes, C.: Low-rank bilinear pooling for fine-grained
  classification. In: Proceedings of the IEEE conference on computer vision and
  pattern recognition. pp. 365--374 (2017)

\bibitem{cars196}
Krause, J., Stark, M., Deng, J., Fei-Fei, L.: 3d object representations for
  fine-grained categorization. In: Proceedings of the IEEE international
  conference on computer vision workshops. pp. 554--561 (2013)

\bibitem{krizhevsky2012imagenet}
Krizhevsky, A., Sutskever, I., Hinton, G.E.: Imagenet classification with deep
  convolutional neural networks. In: Advances in neural information processing
  systems. pp. 1097--1105 (2012)

\bibitem{bim}
Kurakin, A., Goodfellow, I., Bengio, S.: Adversarial examples in the physical
  world. arXiv preprint arXiv:1607.02533  (2016)

\bibitem{long2015fully}
Long, J., Shelhamer, E., Darrell, T.: Fully convolutional networks for semantic
  segmentation. In: Proceedings of the IEEE conference on computer vision and
  pattern recognition. pp. 3431--3440 (2015)

\bibitem{madry2017towards}
Madry, A., Makelov, A., Schmidt, L., Tsipras, D., Vladu, A.: Towards deep
  learning models resistant to adversarial attacks. arXiv preprint
  arXiv:1706.06083  (2017)

\bibitem{mao2019metric}
Mao, C., Zhong, Z., Yang, J., Vondrick, C., Ray, B.: Metric learning for
  adversarial robustness. In: Advances in Neural Information Processing
  Systems. pp. 478--489 (2019)

\bibitem{mustafa2019adversarial}
Mustafa, A., Khan, S., Hayat, M., Goecke, R., Shen, J., Shao, L.: Adversarial
  defense by restricting the hidden space of deep neural networks. In:
  Proceedings of the IEEE International Conference on Computer Vision. pp.
  3385--3394 (2019)

\bibitem{nakka2018deep}
Nakka, K.K., Salzmann, M.: Deep attentional structured representation learning
  for visual recognition. arXiv preprint arXiv:1805.05389  (2018)

\bibitem{narodytska2017simple}
Narodytska, N., Kasiviswanathan, S.: Simple black-box adversarial attacks on
  deep neural networks. In: 2017 IEEE Conference on Computer Vision and Pattern
  Recognition Workshops (CVPRW). pp. 1310--1318. IEEE (2017)

\bibitem{nguyen2016synthesizing}
Nguyen, A., Dosovitskiy, A., Yosinski, J., Brox, T., Clune, J.: Synthesizing
  the preferred inputs for neurons in neural networks via deep generator
  networks. In: Advances in neural information processing systems. pp.
  3387--3395 (2016)

\bibitem{deepfool}
Nguyen, A., Yosinski, J., Clune, J.: Deep neural networks are easily fooled:
  High confidence predictions for unrecognizable images. In: Proceedings of the
  IEEE conference on computer vision and pattern recognition. pp. 427--436
  (2015)

\bibitem{JSMA}
Papernot, N., McDaniel, P., Jha, S., Fredrikson, M., Celik, Z.B., Swami, A.:
  The limitations of deep learning in adversarial settings. In: 2016 IEEE
  European symposium on security and privacy (EuroS\&P). pp. 372--387. IEEE
  (2016)

\bibitem{papernot2016distillation}
Papernot, N., McDaniel, P., Wu, X., Jha, S., Swami, A.: Distillation as a
  defense to adversarial perturbations against deep neural networks. In: 2016
  IEEE Symposium on Security and Privacy (SP). pp. 582--597. IEEE (2016)

\bibitem{ross2018improving}
Ross, A.S., Doshi-Velez, F.: Improving the adversarial robustness and
  interpretability of deep neural networks by regularizing their input
  gradients. In: Thirty-second AAAI conference on artificial intelligence
  (2018)

\bibitem{samangouei2018defense}
Samangouei, P., Kabkab, M., Chellappa, R.: Defense-gan: Protecting classifiers
  against adversarial attacks using generative models. arXiv preprint
  arXiv:1805.06605  (2018)

\bibitem{selvaraju2017grad}
Selvaraju, R.R., Cogswell, M., Das, A., Vedantam, R., Parikh, D., Batra, D.:
  Grad-cam: Visual explanations from deep networks via gradient-based
  localization. In: Proceedings of the IEEE international conference on
  computer vision. pp. 618--626 (2017)

\bibitem{shafahi2019adversarial}
Shafahi, A., Najibi, M., Ghiasi, M.A., Xu, Z., Dickerson, J., Studer, C.,
  Davis, L.S., Taylor, G., Goldstein, T.: Adversarial training for free! In:
  Advances in Neural Information Processing Systems. pp. 3353--3364 (2019)

\bibitem{vgg16}
Simonyan, K., Zisserman, A.: Very deep convolutional networks for large-scale
  image recognition. arXiv preprint arXiv:1409.1556  (2014)

\bibitem{song2017pixeldefend}
Song, Y., Kim, T., Nowozin, S., Ermon, S., Kushman, N.: Pixeldefend: Leveraging
  generative models to understand and defend against adversarial examples.
  arXiv preprint arXiv:1710.10766  (2017)

\bibitem{su2019one}
Su, J., Vargas, D.V., Sakurai, K.: One pixel attack for fooling deep neural
  networks. IEEE Transactions on Evolutionary Computation  \textbf{23}(5),
  828--841 (2019)

\bibitem{szegedy2013intriguing}
Szegedy, C., Zaremba, W., Sutskever, I., Bruna, J., Erhan, D., Goodfellow, I.,
  Fergus, R.: Intriguing properties of neural networks. arXiv preprint
  arXiv:1312.6199  (2013)

\bibitem{tramer2017ensemble}
Tram{\`e}r, F., Kurakin, A., Papernot, N., Goodfellow, I., Boneh, D., McDaniel,
  P.: Ensemble adversarial training: Attacks and defenses. arXiv preprint
  arXiv:1705.07204  (2017)

\bibitem{tsipras2018robustness}
Tsipras, D., Santurkar, S., Engstrom, L., Turner, A., Madry, A.: Robustness may
  be at odds with accuracy. arXiv preprint arXiv:1805.12152  (2018)

\bibitem{cub200}
Wah, C., Branson, S., Welinder, P., Perona, P., Belongie, S.: The caltech-ucsd
  birds-200-2011 dataset  (2011)

\bibitem{wang2016learnable}
Wang, Z., Li, H., Ouyang, W., Wang, X.: Learnable histogram: Statistical
  context features for deep neural networks. In: European Conference on
  Computer Vision. pp. 246--262. Springer (2016)

\bibitem{wong2020fast}
Wong, E., Rice, L., Kolter, J.Z.: Fast is better than free: Revisiting
  adversarial training. arXiv preprint arXiv:2001.03994  (2020)

\bibitem{xie2017mitigating}
Xie, C., Wang, J., Zhang, Z., Ren, Z., Yuille, A.: Mitigating adversarial
  effects through randomization. arXiv preprint arXiv:1711.01991  (2017)

\bibitem{xie2019feature}
Xie, C., Wu, Y., Maaten, L.v.d., Yuille, A.L., He, K.: Feature denoising for
  improving adversarial robustness. In: Proceedings of the IEEE Conference on
  Computer Vision and Pattern Recognition. pp. 501--509 (2019)

\bibitem{xu2018structured}
Xu, K., Liu, S., Zhao, P., Chen, P.Y., Zhang, H., Fan, Q., Erdogmus, D., Wang,
  Y., Lin, X.: Structured adversarial attack: Towards general implementation
  and better interpretability. arXiv preprint arXiv:1808.01664  (2018)

\bibitem{yu2018statistically}
Yu, K., Salzmann, M.: Statistically-motivated second-order pooling. In:
  Proceedings of the European Conference on Computer Vision (ECCV). pp.
  600--616 (2018)

\bibitem{zeiler2014visualizing}
Zeiler, M.D., Fergus, R.: Visualizing and understanding convolutional networks.
  In: European conference on computer vision. pp. 818--833. Springer (2014)

\bibitem{zhang2015cross}
Zhang, C., Li, H., Wang, X., Yang, X.: Cross-scene crowd counting via deep
  convolutional neural networks. In: Proceedings of the IEEE conference on
  computer vision and pattern recognition. pp. 833--841 (2015)

\bibitem{zhang2019theoretically}
Zhang, H., Yu, Y., Jiao, J., Xing, E.P., Ghaoui, L.E., Jordan, M.I.:
  Theoretically principled trade-off between robustness and accuracy. arXiv
  preprint arXiv:1901.08573  (2019)

\bibitem{zhou2016learning}
Zhou, B., Khosla, A., Lapedriza, A., Oliva, A., Torralba, A.: Learning deep
  features for discriminative localization. In: Proceedings of the IEEE
  conference on computer vision and pattern recognition. pp. 2921--2929 (2016)

\end{thebibliography}
	\clearpage

\cutsectionup

\section{Qualitative Results on CUB}
In this section, we provide additional qualitative results on CUB200. In particular, we visualize the learned prototypes, and analyze the classification results by computing the similarity of the samples with the learned prototypes.\\

\noindent\textbf{Visualization of the learned prototypes.}  In Figure~\ref{fig:protobirds}, we show the activation  heat maps of the prototypes on the source images to which they were projected for our VGG-16 model.  Our method yields  fine-grained prototypes that either focus on a small discriminative region or activate the complete non-discriminative region.\\


\providecommand{\localwidth}{}
\renewcommand{\localwidth}{\linewidth}

\providecommand{\localheight}{}
\renewcommand{\localheight}{8cm}

\begin{figure}[b]
	\footnotesize
	\centering
      \includegraphics[width=\textwidth]{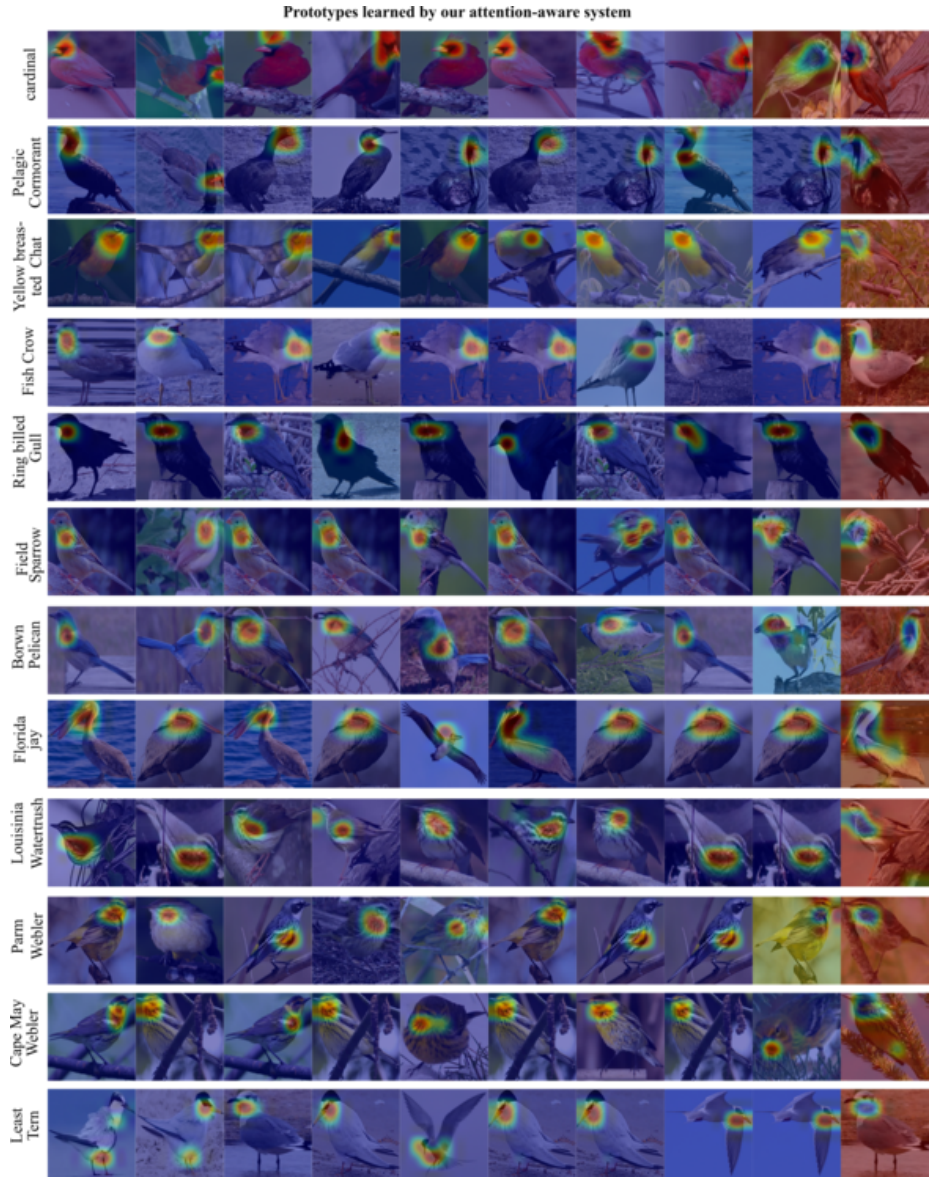}
	\caption{{\bf Visualization of the prototypes learned  with our approach on CUB.} Our formulation yields prototypes that  are fine-grained and representative of the specific class in the images. }
	\label{fig:protobirds}
\end{figure}

\noindent\textbf{Nearest samples of the learned prototypes.}  In Figure~\ref{fig:protobirdsnearest}, we show the prototypes and their nearest training images for CUB 200 with VGG-16. \kn{Similarly, in Figure~\ref{fig:protobirdsnearest_test}, we show the prototypes and their nearest test images for CUB 200 with VGG-16. }
In most cases, the discriminative prototypes activate the same semantic part in all images corresponding to the same class.\\

\noindent\textbf{Nearest prototypes for a clean image.}  In Figure~\ref{fig:protobirdsactivations_clean}, we show, for a given clean test image, the top few highest activated prototypes with VGG-16. We observe that the most activated prototypes focus on salient and discriminative regions, with no influence from the background regions.\\

\noindent\textbf{Nearest prototypes for an adversarial image.}  In Figure~\ref{fig:protobirdsactivations_adv}, we show the top few highest activated codewords for unsuccessful adversarial samples that retain the predicted label even after the attack. Note that, under attack, the similarity scores of the top activated prototypes decrease, but, thanks to the large separation between the  prototypes, the discriminative features do not cross over to other prototypes. 

\comment{Therefore, our system helps to interpret the classification decisions, and also can potentially  detect an adversarial attack by leveraging the prototype similarity scores.
}

\MS{Can we also show the case of successful attacks? In particular, can we interpret these results?}
\KN{ It looks like there are adversarial examples with high similarity scores. }


\providecommand{\localwidth}{}
\renewcommand{\localwidth}{\linewidth}

\providecommand{\localheight}{}
\renewcommand{\localheight}{8cm}

\begin{figure}[b]
	\footnotesize
	\centering
      \includegraphics[width=\textwidth]{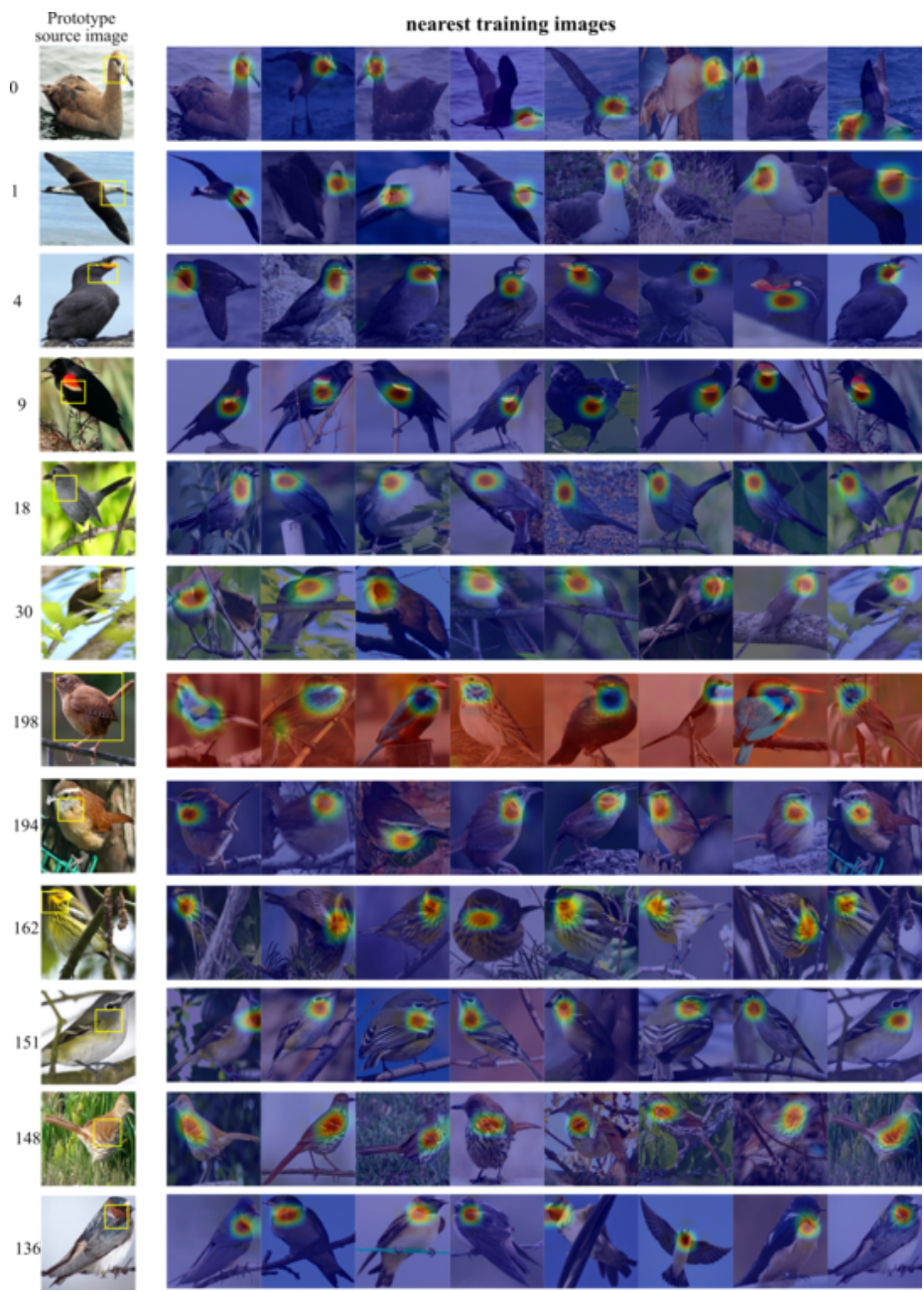}
	\caption{ \textbf{Visualization of the nearest \emph{train} samples for each learned prototypes with our approach on CUB with VGG-16.} All prototypes activate semantically meaningful parts and mostly from the images corresponding to their own label.  }
	\label{fig:protobirdsnearest}
\end{figure}
 

\providecommand{\localwidth}{}
\renewcommand{\localwidth}{\linewidth}

\providecommand{\localheight}{}
\renewcommand{\localheight}{8cm}

\begin{figure}[b]
	\footnotesize
	\centering
      \includegraphics[width=\textwidth]{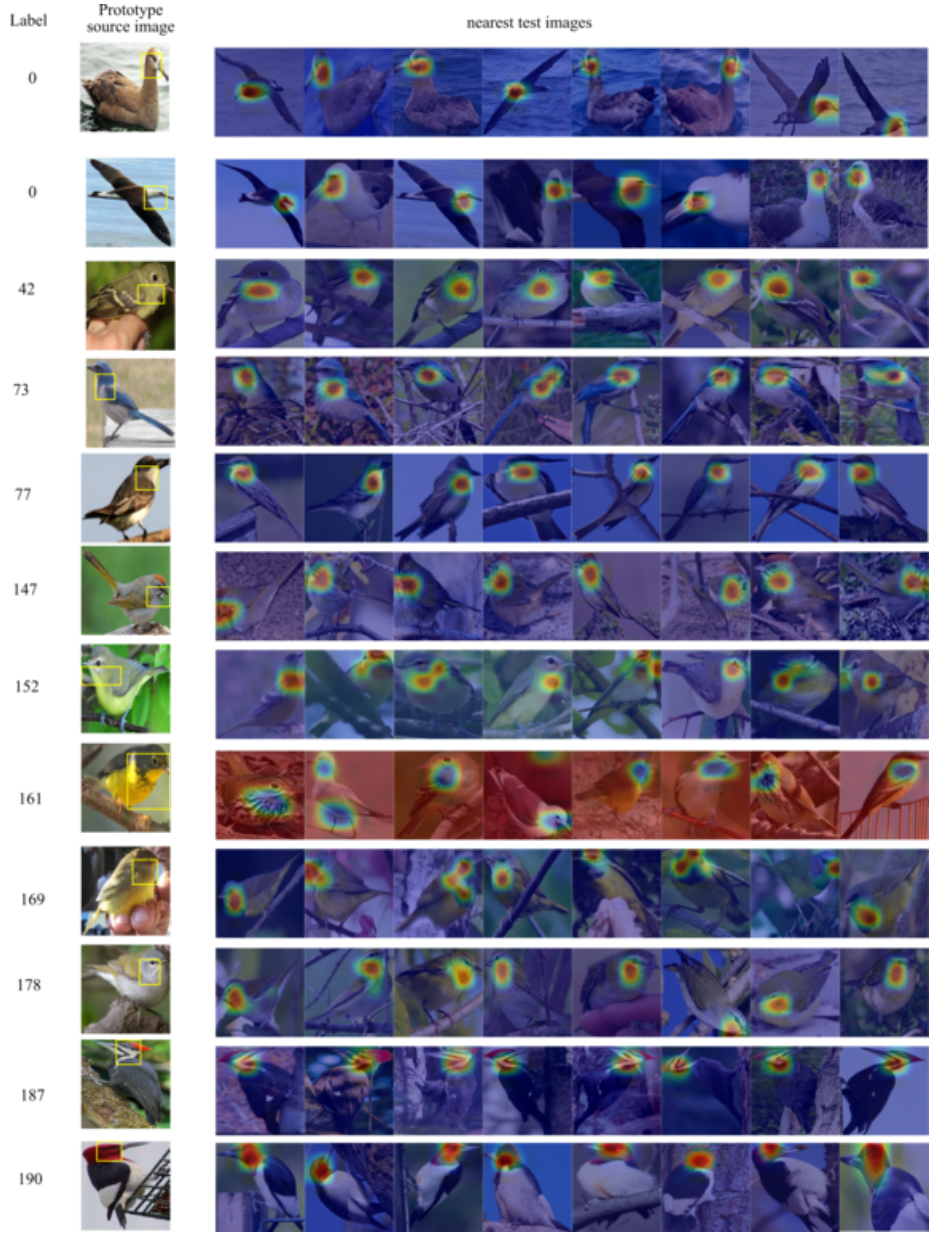}
	\caption{{\bf Visualization of the nearest \emph{test} samples for each learned prototypes with our approach on CUB with VGG-16.} All prototypes activate semantically meaningful parts and mostly from the images corresponding to their own label.  }
	\label{fig:protobirdsnearest_test}
\end{figure}
 

\providecommand{\localwidth}{}
\renewcommand{\localwidth}{\linewidth}

\providecommand{\localheight}{}
\renewcommand{\localheight}{8cm}

\begin{figure}[b]
	\footnotesize
	\centering
      \includegraphics[width=\textwidth]{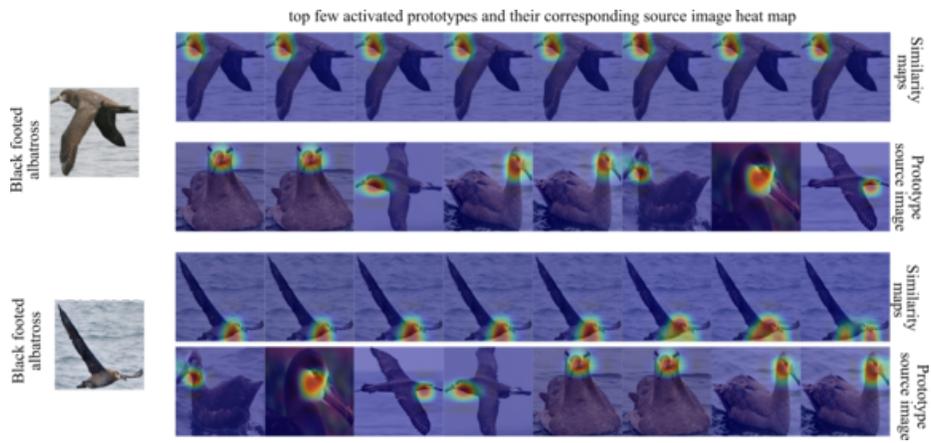}
	\caption{{\bf Visualization of the top activated prototypes for a given test image on CUB}. The top prototypes activate semantically meaningful regions and discard the background areas. }
	\label{fig:protobirdsactivations_clean}
\end{figure}


\providecommand{\localwidth}{}
\renewcommand{\localwidth}{\linewidth}

\providecommand{\localheight}{}
\renewcommand{\localheight}{8cm}

\begin{figure}[b]
	\footnotesize
	\centering
      \includegraphics[width=\textwidth]{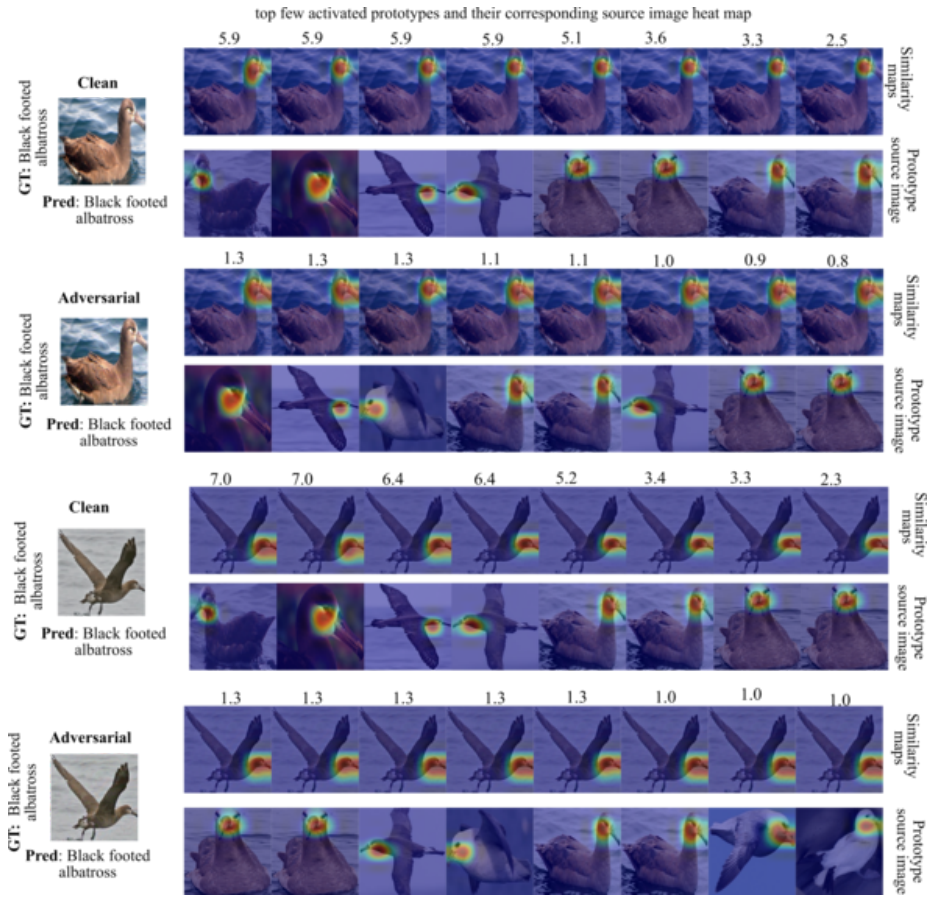}
	\caption{{\bf Visualization of the top activated prototypes for a given clean and adversarial image pair from the CUB test data}. The top prototypes corresponds to the true label, even after attack. Moreover, we observe that the similarity score for each prototype decreases, but the attack remains unsuccessful thanks to the large separation between the discriminative prototypes.}
	\label{fig:protobirdsactivations_adv}
\end{figure}

\section{Results on Stanford Cars}
\cutsectiondown

In Table~\ref{tab:carsAT}, we report the robustness of fast adversarial training~\cite{wong2020fast} with our discriminative feature separation approach. Our approach, \textbf{Ours-FR+AT}, performs better than the baseline  ProtoPNet+AT~\cite{chen2019looks} in all cases. Note that, for multi-step iterative attacks, \textbf{Ours-A+AT} performs better than AP+AT, while they achieve comparable performance for single-step attacks.

\begin{table}[h]
\scriptsize
  \centering
 \setlength\tabcolsep{1pt}
  \begin{tabular}{llllllllllllllll}
    \midrule
      Attacks  & Clean & FGSM& FGSM &  BIM  & BIM  & PGD  &  PGD  &  MIM & MIM & BB-V & BB-D\\    
Steps,$\epsilon$) &  & (1,2) & (1,8) & (10,2)& (10,8) & (10,2) & (10,8)& (10,2) & (10,8) & (10,2) & (10,8)  \\ 
   
     \midrule
    \multicolumn{10}{c}{\textbf{VGG16}} \\
    \midrule
    
    {AP+AT}~\cite{girdhar2017attentional} &\textbf{86.2\%}&\textbf{81.1\%}& \textbf{63.6\%}&\textbf{78.9\%}&53.8\%&\textbf{78.7\%}&50.8\%&\textbf{78.7\%}&55.1\% & \textbf{85.1\%} & \textbf{85.9\%}\\
    {ProtoPNet+AT}~\cite{chen2019looks}&64.4\%&53.7\%&31.9\%&48.9\%&16.5\%&48.2\%&13.4\%&49.2\%&18.2\% & 63.8\%  & 64.2\%\\
   \textbf{Ours-A+AT}&{84.8\%}&79.8\%&63.3\%&77.0\%&54.6\%&76.6\%&51.1\%&77.1\%&\textbf{55.8\%} & 84.5\%  & 85.6\%\\
   \textbf{Ours-FR+AT}&83.7\%&76.37\%&62.8\%&73.5\%&\textbf{55.0\%}&72.6\%&\textbf{51.9\%}&{73.8\%}&{55.4\%} & 80.8\% & 82.0\%\\
   \midrule
   \multicolumn{10}{c}{\textbf{VGG19}} \\
   \midrule
   {AP+AT}~\cite{girdhar2017attentional} & 0\% &0\% &0\% &0\% &0\% &0\% &0\% &0\% &0\% &0\% &0\% & \\
   {ProtoPNet+AT}~\cite{chen2019looks}& 0\% &0\% &0\% &0\% &0\% &0\% &0\% &0\% &0\% &0\% &0\% \\
   \textbf{Ours-A+AT}&\textbf{87.3\%} &\textbf{80.29\%} &\textbf{67.1\%}&\textbf{78.4\%}&\textbf{60.15\%}&\textbf{78.2\%}&\textbf{58.2\%}&\textbf{78.6\%}&\textbf{61.3\%}& \textbf{86.5\%} & \textbf{87.3\%}\\   \textbf{Ours-FR+AT}&84.6\%&{79.6\%}&{66.9\%}&{77.7\%}&{58.6\%}&76.5\%&{55.6\%}&{77.8\%}&{59.1\%} & 83.7\% & 84.5\%\\
   \midrule
   \multicolumn{10}{c}{\textbf{ResNet-34}} \\
   \midrule
   
   {AP+AT}~\cite{girdhar2017attentional} &84.3\%&80.3\%&\textbf{67.0\%}&78.7\%&\textbf{55.3\%}&78.6\%&\textbf{51.9\%}&78.7\%&\textbf{56.8\%}&84.1\% & 84.3\%\\
   {ProtoPNet+AT}~\cite{chen2019looks}&78.1\% &70.4\%&53.5\%&66.3\%&37.9\%&65.7\%&33.0\%&66.8\%&39.1\%& 77.6\% & 78.6\%\\
   \textbf{Ours-A+AT}&\textbf{85.1\%}&\textbf{80.8\%}&{65.9\%}&\textbf{79.0\%}&{54.5\%}&\textbf{78.8\%}&{49.9\%}&\textbf{79.1\%}&56.5\% & \textbf{84.7\%} & \textbf{85.1\%} \\
   \textbf{Ours-FR+AT} &  82.6\% & 77.9\% & 63.8\%  & 75.91\% & 49.97\% & 76.28 & 48.1\% & 76.1\% & 52.4\% & 83.0\% &83.2\%\\
   
   \bottomrule
  \end{tabular}
  \vspace{1mm}
\caption{{\small Classification accuracy of different robust networks with $\ell_{\infty}$ based  attacks on Cars196. The best result of each column and each backbone is shown in \textbf{bold}. The last two columns correspond to black-box attacks. Note that, AP+AT, ProtPNet+AT did not converge for VGG-19.}}
  \label{tab:carsAT}
 \vspace{-7mm}
\end{table}

\noindent\textbf{Visualization of the learned prototypes.}  In Figure~\ref{fig:protocars}, we show the activation  heat maps of the prototypes on the source images to which they were projected for our VGG-16 model.  Our method yields  fine-grained prototypes that either focus on a small discriminative region or activate the complete non-discriminative region.\\

\noindent\textbf{Nearest samples of the learned prototypes.}  In Figure~\ref{fig:protobirdsnearest_cars}, we show the prototypes and their nearest training images for Cars 196 with VGG-16. \kn{Similarly, in Figure~\ref{fig:protocarsnearest_test}, we show the prototypes and their nearest test images for Cars 196 with VGG-16. }
In most cases, the discriminative prototypes activate the same semantic part in all images corresponding to the same class.\\

\noindent\textbf{Nearest prototypes for an adversarial image.}  In Figure~\ref{fig:protobirdsactivations_adv_cars}, we show the top few highest activated codewords for unsuccessful adversarial samples that retain the predicted label even after the attack. Note that, under attack, the similarity scores of the top activated prototypes decrease, but, thanks to the large separation between the  prototypes, the discriminative features do not cross over to other prototypes.


\providecommand{\localwidth}{}
\renewcommand{\localwidth}{\linewidth}

\providecommand{\localheight}{}
\renewcommand{\localheight}{8cm}

\begin{figure}[b]
	\footnotesize
	\centering
      \includegraphics[width=\textwidth]{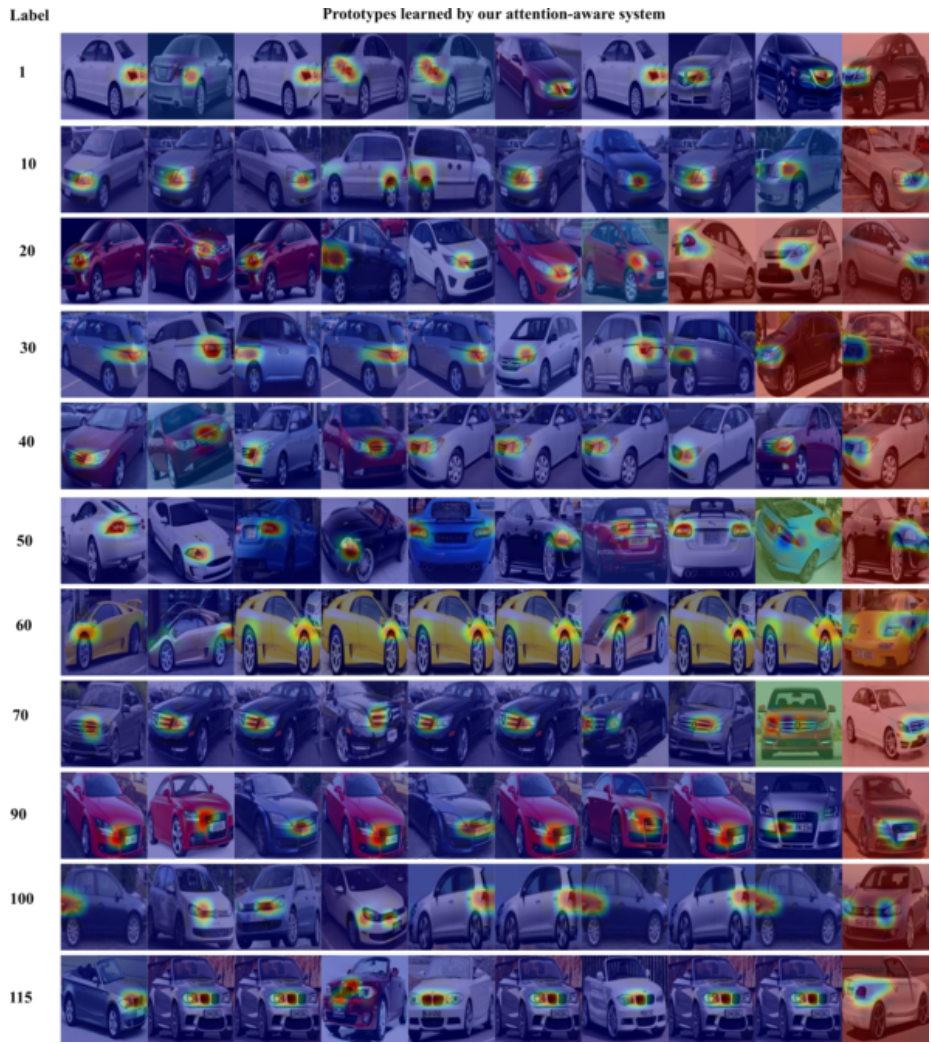}
	\caption{{\bf Visualization of the prototypes learned  with our approach on Cars-196.} Our formulation yields prototypes that  are fine-grained and representative of the specific class in the images. }
	\label{fig:protocars}
\end{figure}
 

\providecommand{\localwidth}{}
\renewcommand{\localwidth}{\linewidth}

\providecommand{\localheight}{}
\renewcommand{\localheight}{8cm}

\begin{figure}[b]
	\footnotesize
	\centering
      \includegraphics[width=\textwidth]{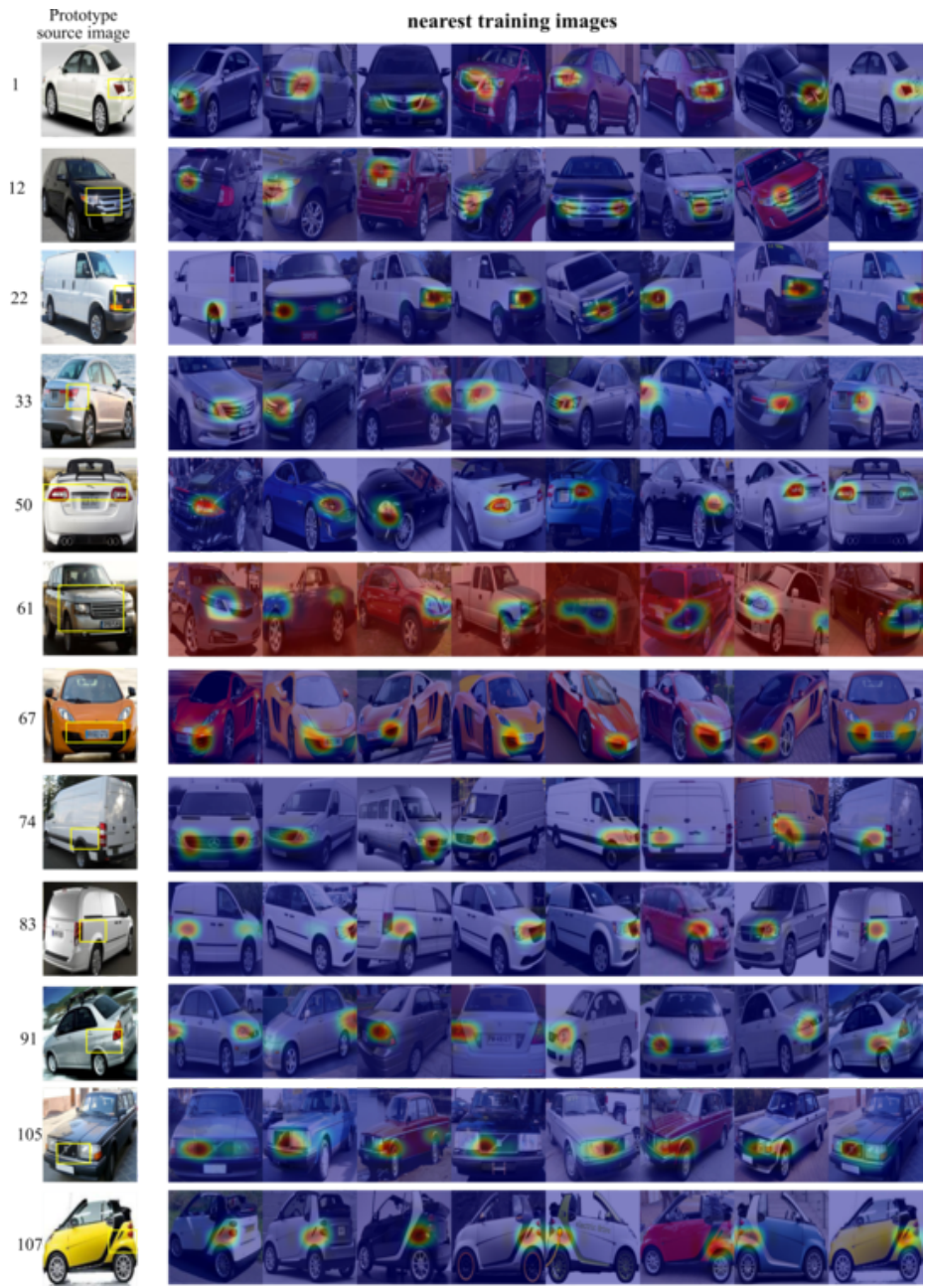}
	\caption{ \textbf{Visualization of the nearest \emph{train} samples for each learned prototypes with our approach on Cars-196 with VGG-16.} All prototypes activate semantically meaningful parts and mostly from the images corresponding to their own label.  }
	\label{fig:protobirdsnearest_cars}
\end{figure}
 

\providecommand{\localwidth}{}
\renewcommand{\localwidth}{\linewidth}

\providecommand{\localheight}{}
\renewcommand{\localheight}{8cm}

\begin{figure}[b]
	\footnotesize
	\centering
      \includegraphics[width=\textwidth]{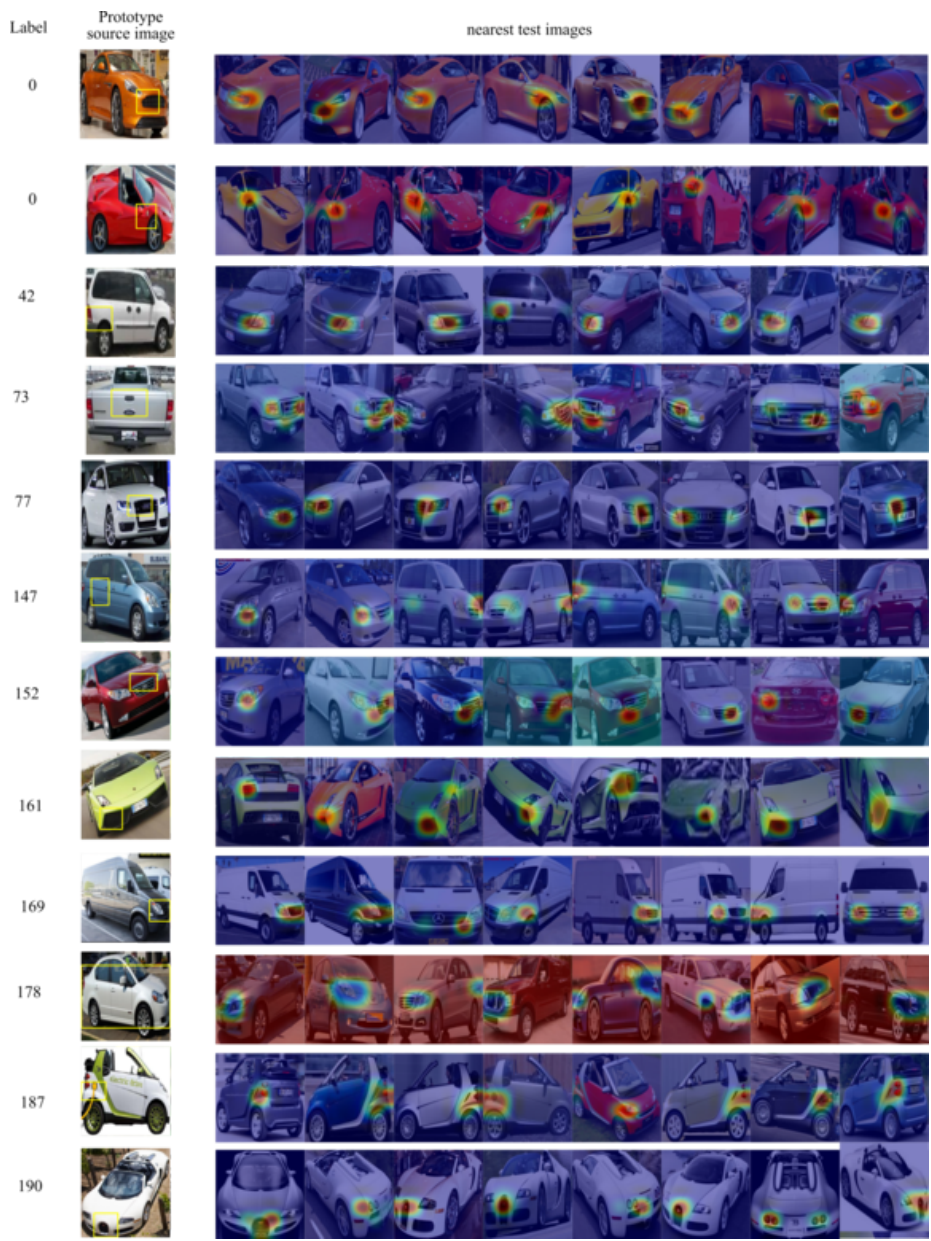}
	\caption{{\bf Visualization of the nearest \emph{test} samples for each learned prototypes with our approach on Cars 196 with VGG-16.} All prototypes activate semantically meaningful parts and mostly from the images corresponding to their own label.  }
	\label{fig:protocarsnearest_test}
\end{figure}
 

\providecommand{\localwidth}{}
\renewcommand{\localwidth}{\linewidth}

\providecommand{\localheight}{}
\renewcommand{\localheight}{8cm}

\begin{figure}[b]
	\footnotesize
	\centering
      \includegraphics[width=\textwidth]{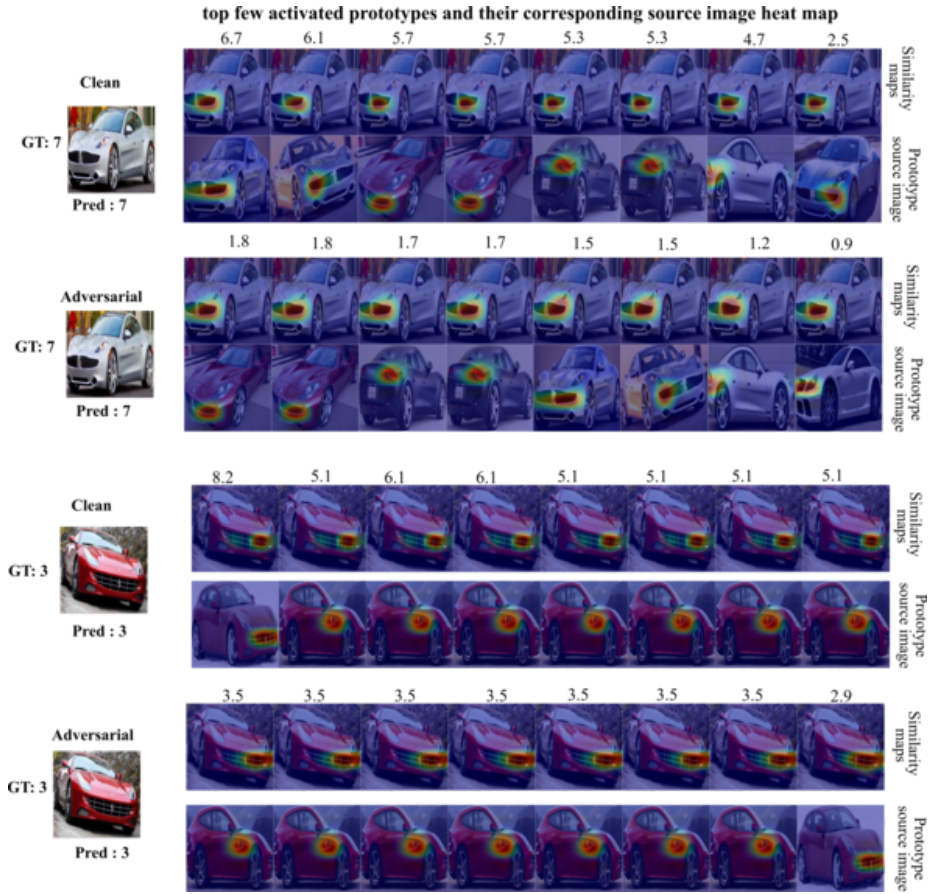}
	\caption{{\bf Visualization of the top activated prototypes for a given clean and adversarial image pair from the Cars-196 test data}. The top prototypes corresponds to the true label, even after attack. Moreover, we observe that the similarity score for each prototype decreases, but the attack remains unsuccessful thanks to the large separation between the discriminative prototypes.}
	\label{fig:protobirdsactivations_adv_cars}
\end{figure}

\end{document}